\documentclass{article} \usepackage{iclr2024_preprint,times}
\iclrfinalcopy

\usepackage{amsmath,amsfonts,bm}

\def\eqref#1{equation~\ref{#1}}

\def\1{\bm{1}}

\DeclareMathAlphabet{\mathsfit}{\encodingdefault}{\sfdefault}{m}{sl}
\SetMathAlphabet{\mathsfit}{bold}{\encodingdefault}{\sfdefault}{bx}{n}

\usepackage{hyperref}
\usepackage{url}

\usepackage{subcaption}
\usepackage{multirow}
\usepackage{makecell}
\usepackage{booktabs}

\usepackage{comment}
\usepackage[noend]{algpseudocode}
\usepackage{enumitem}
\usepackage[font={footnotesize}]{caption}

\setlist[enumerate]{
                    itemindent=1em,
                    leftmargin=0em, }
                    
\title{\modelname: Learning Continuous Image Representation for 
Spatial-Spectral
Super-Resolution}

\author{
Gengchen Mai$^{1,2}$, Ni Lao$^3$, Weiwei Sun$^4$, Yuchi Ma$^5$, Jiaming Song$^2$, Chenlin Meng$^2$, \\
       \textbf{Hongxu Ma}$^3$, \textbf{Jinmeng Rao}$^3$, \textbf{Ziyuan Li}$^6$ \& \textbf{Stefano Ermon}$^2$ \\
$^1$Spatially Explicit Artificial Intelligence Lab, Department of Geography, University of Georgia \\\texttt{gengchen.mai25@uga.edu} \\
$^2$Department of Computer Science, Stanford University\\\texttt{\{maigch,tsong,chenlin,ermon\}@cs.stanford.edu} \\
$^3$Alphabet Inc. \\\texttt{\{nlao,hxma,jinmengrao\}@google.com} \\
$^4$Department of Computer Science, University of British
Columbia \\\texttt{weiweis@cs.ubc.ca} \\
$^5$Department of Earth System Science, Stanford University \\\texttt{yuchima@stanford.edu} \\
$^6$School of Business, University of Connecticut \\\texttt{ziyuan.2.li@uconn.edu} \\
}

\usepackage{amsthm}
\usepackage{amsmath}
\usepackage{xspace}
\usepackage{color}
\usepackage{pgfplots, pgfplotstable}
\usepackage[colorinlistoftodos]{todonotes}

\newcommand{\w}[0]{0.495}

\def\w0495{0.395}

\usepackage[normalem]{ulem}
\useunder{\uline}{\ul}{}

\renewcommand{\vec}[1]{\boldsymbol{\mathbf{#1}}}

\def\bm{\vec{m}}

\def\Real{\mathbb{R}}

\newcommand{\nscale}{T}
\newcommand{\scalek}{t}

\newcommand{\scalefac}{g}

\newcommand{\maxscale}{\lambda_{max}}
\newcommand{\minscale}{\lambda_{min}}

\newcommand{\modelfullname}{Spatial-Spectral Implicit Function}
\newcommand{\modelname}{SSIF}

\newcommand{\hlr}{h}
\newcommand{\wlr}{w}
\newcommand{\clr}{c}
\newcommand{\hhr}{H}
\newcommand{\whr}{W}
\newcommand{\chr}{C}

\newcommand{\cmin}{\chr_{min}}
\newcommand{\cmax}{\chr_{max}}

\newcommand{\band}{b}

\newcommand{\img}{\mathbf{I}}
\newcommand{\HRHSI}{hr-h}
\newcommand{\HRMSI}{hr-m}

\newcommand{\LRMSI}{lr-m}
\newcommand{\imgHRHSI}{\img_{\HRHSI}}
\newcommand{\imgHRHSIMAX}{\img^{\prime}_{\HRHSI}}
\newcommand{\imgHRMSI}{\img_{\HRMSI}}

\newcommand{\imgLRMSI}{\img_{\LRMSI}}

\newcommand{\uniformdist}{Uni}

\newcommand{\scalespa}{p}
\newcommand{\scalespaMIN}{p_{min}}
\newcommand{\scalespaMAX}{p_{max}}

\newcommand{\pt}{\mathbf{x}}
\newcommand{\ptval}{\mathbf{s}}
\newcommand{\wave}{\lambda}

\newcommand{\respfun}{\rho}
\newcommand{\radafunc}{\gamma}

\newcommand{\ptid}{l}
\newcommand{\ptHRHSI}[1]{\pt_{\ptid #1}}

\newcommand{\ptvalHRHSI}[1]{\ptval_{\HRHSI #1}}
\newcommand{\wavevecHRHSI}{\Lambda}

\newcommand{\ptvalgt}[1]{\ptval^{\prime}_{#1}}

\newcommand{\wavemat}{\Lambda}
\newcommand{\wavematHRHSI}{\wavemat_{\HRHSI}}

\newcommand{\encimg}{E_{I}}
\newcommand{\encimgcdim}{d_{I}}

\newcommand{\decsif}{F_{\pt}}

\newcommand{\encband}{E_{\wave}}
\newcommand{\decdim}{d}
\newcommand{\sifhid}{\mathbf{h}_{\ptid}}

\newcommand{\embband}[1]{\mathbf{b}_{ #1}}

\newcommand{\waveidx}{k}
\newcommand{\numsampwave}{K}
\newcommand{\param}{{}\cdot{}}

\newcommand{\decband}{D_{\pt,\wave}}

\newcommand{\srmodel}{H_{sr}}

\newcommand{\ssif}{G_{x,\wave}}

\newcommand{\feaLRMSI}{\mathbf{S}_{\LRMSI}}

\newcommand{\dataset}{\mathcal{D}}
\newcommand{\lossfun}{\mathcal{L}}

\newcommand{\psenc}{\Psi}

\newcommand{\specmlp}{MLP}

\newcommand{\cardprod}{\odot}

\newcommand{\liifmlphiddim}{h_{LIIF}}
\newcommand{\ssifmlphiddim}{h_{SSIP}} 
\begin{document}

\maketitle

\begin{abstract}
Existing digital sensors capture images at fixed spatial and spectral resolutions (e.g., RGB, multispectral, and hyperspectral images), and each combination requires bespoke machine learning models.
Neural Implicit Functions partially overcome the spatial resolution challenge by representing an image in a resolution-independent way.
However, they still operate at fixed, pre-defined spectral resolutions.
To address this challenge, we propose \modelfullname~ (\modelname), a neural implicit model that represents an image as a function of both continuous pixel coordinates
in the spatial domain and continuous wavelengths in the spectral domain. We empirically demonstrate the effectiveness of \modelname~ on two challenging spatio-spectral super-resolution benchmarks. We observe that \modelname~ consistently outperforms state-of-the-art baselines even when the baselines are allowed to train separate models at each spectral resolution. 
We show that SSIF generalizes well to both unseen spatial resolutions and spectral resolutions. Moreover, \modelname{} can generate high-resolution images that improve the performance of downstream tasks (e.g., land use classification) by 1.7\%-7\%.
\end{abstract}

\vspace{-0.3cm}
\section{Introduction}  \label{sec:intro}
\vspace{-0.3cm}

While the physical world is continuous,  most digital sensors (e.g., 
cell phone cameras, multispectral or hyperspectral sensors in satellites) can only capture a discrete representation of continuous signals in both spatial and spectral domains (i.e., with a fixed number of spectral bands, such as red, green, and blue). 
In fact, due to the limited energy of incident photons, fundamental limitations in achievable signal-to-noise ratios (SNR), and time constraints, there is always a trade-off between spatial and spectral resolution \citep{mei2020SepSSJSR,ma2021US3RN}\footnote{Given a fixed overall sensor size and exposure time, higher spatial resolution and higher spectral resolution require the per pixel sensor to be smaller and bigger at the same time, which are contradicting each other.}. High spatial resolution and high spectral resolution can not be achieved at the same time, leading to a variety of 
spatial and spectral resolutions used in practice for different sensors. 
However, ML models are typically bespoke to certain resolutions, and models typically do not generalize to spatial or spectral resolutions they have not been trained on.
This calls for image super-resolution methods.

The goal of image super-resolution (SR) \citep{ledig2017photo,lim2017EDSR,zhang2018RDN,haris2018deep,zhang2020unsupervised,yao2020cross,mei2020SepSSJSR,saharia2021SR3,ma2021US3RN,he2021spatial}
is to increase the spatial or spectral resolution of a given single low-resolution image~\citep{galliani2017learned}.
It has become increasingly important for 
a wide range of tasks including object recognition and tracking \citep{2003pan,2015uzair,2020xiong}, medical image processing \citep{Lu2014MedicalHI,Johnson2007SnapshotHI},  remote sensing \citep{he2021spatial,Bioucas2013,Melgani2004,Zhong2018,Wang2022} and astronomy \citep{2019ehtc}.

Traditionally image SR has been classified into three tasks
according to the input and output image resolutions:\footnote{A related task, Multispectral and Hyperspectral Image Fusion \citep{zhang2020unsupervised,yao2020cross}, takes a high spatial resolution multispectral image and a low spatial resolution hyperspectral image as inputs and generates a high-resolution hyperspectral image. In this paper, we focus on the single image-to-image translation problem and leave this task as the future work.}
Spatial Super-Resolution (spatial SR), 
Spectral Super-Resolution (spectral SR) and 
Spatio-Spectral Super-Resolution (SSSR). 
Spatial SR \citep{zhang2018RCAN,hu2019MetaSR,zhang2020USRNet,niu2020single,wu2021scale,chen2021LIIF,he2021spatial}
focuses on increasing the spatial resolution of the input images (e.g., from $\hlr \times \wlr$ pixels to $\hhr \times \whr$ pixels) while keeping the spectral resolution (\textit{i.e.}, number of spectral bands/channels) unchanged.
In contrast, spectral SR \citep{galliani2017learned,zhang2021implicit} focuses on increasing the spectral resolution of the input images (e.g., from $\clr$ to $\chr$ channels) while keeping the spatial resolution fixed. 
SSSR  \citep{mei2020SepSSJSR,ma2021US3RN} focuses on increasing both the spatial and spectral resolution of the input images. 
Here, $\hlr, \wlr$ (or $\hhr, \whr$ ) indicates the height and width of the low-resolution, LR, (or high-resolution, HR) images while $\clr$ and $\chr$ indicates the number of bands/channels of the low/high spectral resolution images. 
For video signal, SR can also be done along the time dimension, but we don't consider it here and leave it as future work.

\begin{figure*}[t!]
	\centering 
	\includegraphics[width=0.75\textwidth]{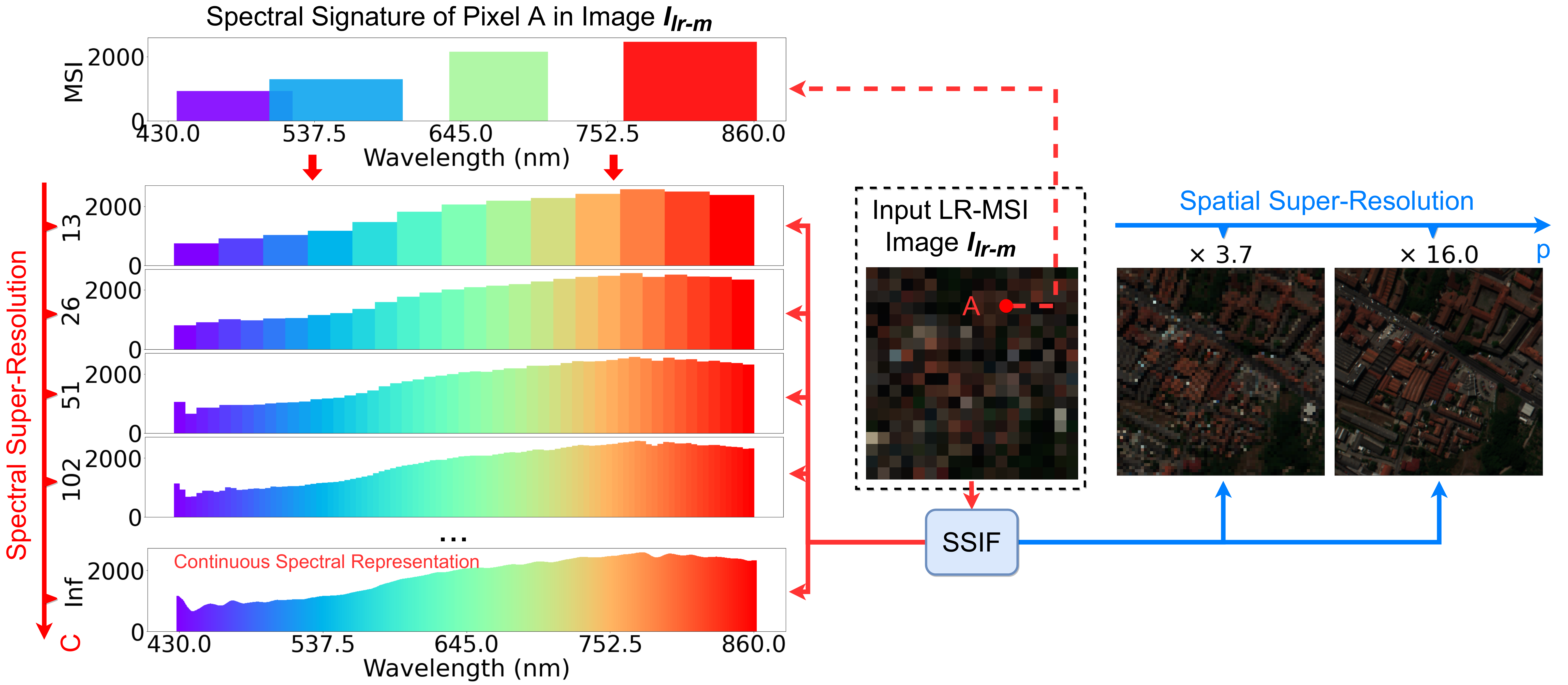}\vspace*{-0.2cm}
	\caption{
	\modelfullname~ (\modelname).
Given an input low-resolution multispectral (LR-MSI) image,
\modelname~ can perform both spatial (blue arrows) and spectral (red arrows) super-resolution simultaneously (illustrated with a specific pixel A). 
Unlike all the other neural implicit functions \modelname~  can generate images with any number of bands
 including ``Inf" -- a continuous function.
}
	\label{fig:ssif_cont}
	\vspace*{-0.15cm}
\end{figure*} 
The diversity in input-output image resolutions (both spatial and spectral) 
significantly increases the complexity of developing deep neural network (DNN)--based SR models. 
Instead of jointly learning representations from images with different spatial and spectral resolutions, most SR research develops separate DNN models for each input-output image resolution pairs with a specific spatial and spectral resolution \citep{lim2017EDSR,zhang2018RDN,ma2021US3RN,mei2020SepSSJSR}.
For example, convolution-based SR models such as RCAN \citep{zhang2018RCAN}, SR3\citep{saharia2021SR3}, SSJSR \citep{mei2020SepSSJSR} and \citep{he2021spatial} need to be trained separately for each input-output image resolution settings\footnote{Figure \ref{fig:moti-1} in Appendix \ref{sec:multitask_illus} illustrates this separate training practice.}. This practice has two limitations: 1) For some SR settings with much less training data, these models can yield suboptimal results or lead to overfitting; 
2) It prevents generalizing trained SR models to unseen spatial/spectral resolutions. 

Inspired by the recent progress in 3D reconstruction with implicit neural representation \citep{park2019deepsdf,mescheder2019occupancy,chen2019learning,sitzmann2020implicit,mildenhall2020nerf}, image neural implicit functions (NIF) \citep{dupont2021coin,chen2021LIIF,yang2021ITSRN,zhang2021implicit} partially overcome the aforementioned problems (especially the second one) by learning a continuous function that maps an arbitrary pixel spatial coordinate to the corresponding visual signal value; 
so in principle, they can generate images at any spatial resolution. 
For example, 
LIIF \citep{chen2021LIIF}
is capable of generating images at any arbitrary resolution in the spatial domain. We call them \textit{Spatial Implicit Functions (SIF)}.
However, all current implicit function representations only focus on generalization in the spatial domain, and each SIF model is trained separately to target a specific spectral resolution (i.e., a fixed number of spectral bands).

In this work, we propose \modelfullname~ ($\modelname$), which generalizes the idea of neural implicit representations to the spectral domain.
$\modelname$ represents an image as a continuous function on both pixel spatial coordinates in the spatial domain and wavelengths in the spectral domain.
As shown in Figure \ref{fig:ssif_cont}, given an input low-resolution multispectral (or RGB) 
image 
, a single $\modelname$ model can generate images with different spatial resolutions and spectral resolutions.
Note that extending the idea of implicit representations to the spectral domain is a non-trivial task. LIIF and other NIF models have an equal distance assumption in the spatial domain, meaning that pixels in the target HR image are assumed to be equally spaced. However, this equal distance assumption does not necessarily hold in the spectral domain. For many RGB or multispectral images, each band may have different spectral widths, i.e.,  wavelength intervals of different lengths. Moreover, the wavelength intervals of different bands may overlap with each other. The "Spectral Signature of Pixel A" of the image 
$\imgLRMSI$ 
in Figure \ref{fig:ssif_cont} shows one example of such cases.
To tackle this problem, we predict each spectral band value of each target pixel separately as the integral of the correlation between the pixel's radiance function and the current band's spectral response function over the desired spectral interval. 
Our contributions are as follows:
\begin{enumerate}
\setlength\itemsep{-0.2em}
     \item We propose \modelfullname~ ($\modelname$) which represents an image as a continuous function on both pixel coordinates in the spatial domain and wavelengths in the spectral domain. $\modelname$ can handle SR tasks with different spatial and spectral resolutions simultaneously.
    \item We demonstrate the effectiveness of $\modelname$~ on two challenging spatio-spectral super-resolution benchmarks -- CAVE (the indoor scenes) and Pavia Centre (Hyperspectral Remote Sensing images). We show that \modelname{} consistently outperforms state-of-the-art SR baseline models even when the baselines 
are trained 
    separately at each spectral resolution (and spatial resolution), thus solving an easier task. Moreover, \modelname~ generalizes well to both unseen spatial resolutions and spectral resolutions.
    \item We test the fidelity of the generated high resolution images on the downstream task of land use classification. Compared with the baselines, the images generated by $\modelname$~ have much higher classification accuracy with 1.7\%-7\% performance improvements.
\end{enumerate}

 \vspace{-0.3cm}
\section{Related Work}  \label{sec:related}
\vspace{-0.3cm}

\paragraph{Multispectral and Hyperspectral Image Super-Resolution}
As an ill-posed single image-to-image translation problem, super-resolution (SR) aims at increasing the spatial or spectral resolution of a given image such that it can be used for different downstream tasks. It has been widely used on natural  images\citep{zhang2018RCAN,hu2019MetaSR,zhang2020deep,saharia2021SR3,chen2021LIIF}, screen-shot images \citep{yang2021ITSRN}, omnidirectional images \citep{deng2021lau,yoon2021spheresr}
medical images \citep{isaac2015super}, as well as multispectral \citep{he2021spatial} and hyperspectral remote sensing images\citep{mei2017hyperspectral,ma2021US3RN,mei2020SepSSJSR,wang2022hyperspectral}. It can be classified into three categories: spatial SR, spectral SR, and spatiospectral SR (SSSR). In this work, we focus on the most challenging task, SSSR, which subsumes spatial SR and spectral SR.

\paragraph{Implicit Neural Representation}
Recently, we have witnessed an increasing amount of work using implicit neural representations for different tasks such as image regression \citep{tancik2020fourier} and compression\citep{dupont2021coin,strumpler2021implicit}, 3D shape regression/reconstruction \citep{mescheder2019occupancy,tancik2020fourier,chen2019learning}, 3D shape reconstruction via image synthesis \citep{mildenhall2020nerf}, 3D magnetic resonance imaging (MRI) reconstruction \citep{tancik2020fourier}, 3D protein reconstruction \citep{zhong2020reconstructing}, spatial feature distribution modeling \citep{mai2019space2vec,mai2022review,mai2023sphere2vec}, remote sensing image classification \citep{mai2023csp}, geographic question answering \citep{mai2020se}, and etc.. The core idea is to learn a continuous function that maps spatial coordinates (e.g., pixel coordinates, 3D coordinates, and geographic coordinates) to the corresponding signals (e.g., point cloud intensity, MRI intensity, visual signals, etc.). 
A common setup is to input the spatial coordinates in a deterministic or learnable Fourier feature mapping layer \citep{tancik2020fourier} (consisting of sinusoidal functions with different frequencies), which converts the coordinates into multi-scale features. Then a multi-layer perceptron takes this multi-scale feature as input and whose output is used for downstream tasks. 
In parallel, implicit neural functions (INF) such as LIIF \citep{chen2021LIIF}, ITSRN \citep{yang2021ITSRN}, \cite{zhang2021implicit} are proposed for image super-resolution which map pixel spatial coordinates to the visual signals in the high spatial resolution images. One outstanding advantage is that they can jointly handle SR tasks at an arbitrary spatial scale. 
However, all the existing implicit functions learn continuous image representations in the spatial domain while still operating at fixed, pre-defined spectral resolutions. Our proposed \modelname~ overcomes this problem and generalizes INF to both spatial and spectral domains.

 \vspace{-0.3cm}
\section{Problem Statement} \label{sec:prob_stat}
\vspace{-0.3cm}
The spatial-spectral image super-resolution (SSSR) problem over various spatial and spectral resolutions can be conceptualized as follows. Given an input low spatial/spectral resolution (LR-MSI) image $\imgLRMSI \in \Real^{\hlr \times \wlr \times \clr}$, we want to generate a high spatial/spectral resolution (HR-HSI) image $\imgHRHSI \in \Real^{\hhr \times \whr \times \chr}$. Here, $\hlr, \wlr, \clr$ and $\hhr, \whr, \chr$ are the height, width and channel dimension of image $\imgLRMSI$ and $\imgHRHSI$, and $\hhr > \hlr$,  $\whr > \wlr$,  $\chr > \clr$.
The spatial upsampling scale $\scalespa$ is defined as $\scalespa = \hhr/\hlr = \whr/\wlr$. 
Without loss of generality, let $\wavematHRHSI = [\wavevecHRHSI_{0}^{T}, \wavevecHRHSI_{1}^{T},..., \wavevecHRHSI_{\chr}^{T}] \in \Real^{\chr \times 2}$ be the wavelength interval matrix, which defines the spectral bands in the target HR-HSI image $\imgHRHSI$. Here, $\wavevecHRHSI_{i} = [\wave_{i,s}, \wave_{i,e}] \in \Real^{2}$ is the wavelength interval for the $i$th band of $\imgHRHSI$ where $\wave_{i,s}, \wave_{i,e}$ are the start and end wavelength of this band.
$\wavematHRHSI$ can be used to fully express the spectral resolution of the target HR-HSI image $\imgHRHSI$. In this work, we do not use $\chr/\clr$ to represent the spectral upsampling scale because bands/channels of image $\imgLRMSI$ and $\imgHRHSI$ might not be equally spaced (See Figure \ref{fig:ssif_cont}). 
So $\wavematHRHSI$ is a very flexible representation for the spectral resolution, capable of representing situations when different bands have different spectral widths or their wavelength intervals overlap with each other.
When $\imgHRHSI$ has equally spaced wavelength intervals, such as those of most of the hyperspectral images, we use its band number $\chr$ to represent the spectral scale. 

The spatial-spectral super-resolution (SSSR) can be represented as a function
\begin{align}
\imgHRHSI = \srmodel(\imgLRMSI, \scalespa, \wavematHRHSI)
\label{eq:sr}
\end{align}
where $\srmodel(\cdot)$ takes as input the image $\imgLRMSI$, the desired spatial upsampling scale $\scalespa$, and the target sensor wavelength interval matrix $\wavematHRHSI$, and generates the HR-HSI image $\imgHRHSI \in \Real^{\hhr \times \whr \times \chr}$. 
In other words, we aim at learning \textbf{one single function} $\srmodel(\cdot)$ that can take any input images $\imgLRMSI$ with a fixed spatial and spectral resolution, and generate images $\imgHRHSI$ with diverse spatial and spectral resolutions specified by different $\scalespa$ and $\wavematHRHSI$.

Note that none of the existing SR models can achieve this. Most classic SR models have to learn separate $\srmodel(\cdot)$ for different pairs of $\scalespa$ and $\wavematHRHSI$ such as RCAN \citep{zhang2018RCAN}, SR3\citep{saharia2021SR3}, SSJSR \citep{mei2020SepSSJSR}, \cite{he2021spatial}. 
As for Spatial Implicit Functions (SIF) such as LIIF\citep{chen2021LIIF}, SADN \citep{wu2021SADN}, ITSRN \citep{yang2021ITSRN}, \cite{zhang2021implicit}, they can learn one $\srmodel(\cdot)$ for different $\scalespa$ but with a fixed $\wavematHRHSI$. \vspace{-0.3cm}
\section{\modelfullname} \label{sec:method}
\vspace{-0.3cm}
\subsection{Sensor principles}
To design \modelname, we follow the physical principles of spectral imaging.
Let $\ptval_{\ptid,i}$ be the pixel density value of a pixel $\ptHRHSI{}$ at the spectral band $\band_i$ with wavelength interval $\wavevecHRHSI_{i}$.
It can be computed by an integral of the \textbf{radiance function} $\radafunc_{\img}(\ptHRHSI{}, \wave)$ and 
\textbf{response function} $\respfun_i(\wave)$ of a sensor at band $\band_i$. 
\begin{align}
\ptval_{\ptid,i} = \int_{\wavevecHRHSI_{i}}  \respfun_i(\wave) \radafunc_{\img}(\ptHRHSI{}, \wave) \, \mathrm{d}\wave 
\label{eq:respfunc}
\end{align}
where $\wave$ is wavelength. 
So for each pixel $\ptHRHSI{}$, the radiance function is a neural field that describes the radiance curve as a function of the wavelength. 
Note that unlike recent NeRF where only three discrete wavelength intervals (i.e., RGB) are considered, we aim to learn a \emph{continuous} radiance curve for each pixel.           
The spectral response function \citep{zheng2020coupled} describes the sensitivity of the sensor to different wavelengths and is usually sensor-specific. 
For example, the red sensor in commercial RGB cameras has a strong response (i.e., high pixel density) to red light. 
The spectral response functions of many commercial hyperspectral sensors (e.g., AVIRIS's ROSIS-03\footnote{\url{https://crs.hi.is/?page_id=877}}, EO-1 Hyperion) are very complex due to atmospheric absorption. A common practice adopted by many studies \citep{barry2002eo1,brazile2008toward,cundill2015adjusting,crawford2019radiometric,chi2021spectral} is to approximate the response function of individual spectral bands as a Gaussian distribution or a uniform distribution. In this work, we adopt this practice and show that 
this inductive bias enforced via physical laws improves generalization.

In the following, we will discuss the design of our \modelname~ which allows us to train a single SR model for different $\scalespa$ and $\wavematHRHSI$. 
The whole model architecture of \modelname~ is illustrated in Figure \ref{fig:model_arch}.

\subsection{\modelname~ Architecture}   \label{sec:ssip}
Following  previous SIF works \citep{chen2021LIIF,yang2021ITSRN}, \modelname~ first uses an image encoder $\encimg(\cdot)$ to convert the input image $\imgLRMSI \in \Real^{\hlr \times \wlr \times \clr}$ into a 2D feature map $\feaLRMSI = \encimg(\imgLRMSI) \in \Real^{\hlr \times \wlr \times \encimgcdim}$ which shares the same spatial shape as $\imgLRMSI$ but with a larger channel dimension.
$\encimg(\cdot)$ can be any convolution-based image encoder such as EDSR \citep{lim2017EDSR} or RDN \citep{zhang2018RDN}.

\modelname{} approximates the mathematical integral shown in Equation \ref{eq:respfunc} as a weighted sum over the predicted radiance values of $\numsampwave$ wavelengths $\{\wave_{i,1},\wave_{i,2},...,\wave_{i,\numsampwave}\}$ sampled from a wavelength interval $\wavevecHRHSI_{i} = [\wave_{i,s}, \wave_{i,e}]  \in \wavematHRHSI$
at location $\ptHRHSI{}$ (see Equation \ref{eq:ssif}).
Here, $\respfun_i(\wave_{i,\waveidx})$ is the response function value, i.e., weight, of each wavelength $\wave_{i,\waveidx}$ given the current response function for band $\band_i$.
$\radafunc_{\img}(\ptHRHSI{}, \wave_{i,\waveidx})$ is the radiance value of $\wave_{i,\waveidx}$ at location $\ptHRHSI{}$ which can be computed by a neural implicit function $\ssif$.
Basically, $\ssif$ maps an arbitrary pixel location $\ptHRHSI \in [-1,1] \cardprod [-1, 1]$ of $\imgHRHSI$ and 
a wavelength $\wave_{i,\waveidx} \in \wavevecHRHSI_{i}$
into the radiance value of the target image $\imgHRHSI$ at the corresponding location and wavelength, i.e., $\radafunc_{\img}(\ptHRHSI{}, \wave_{i,\waveidx}) = \ssif(\feaLRMSI, \ptHRHSI{}, \wave_{i,\waveidx})$. Here, $\cardprod$ is the Cartesian product. 
\vspace{-0.1cm}
\begin{align}
\ptval_{\ptid,i} 
= \sum_{\waveidx=1}^{\numsampwave} \respfun_i(\wave_{i,\waveidx}) \radafunc_{\img}(\ptHRHSI{}, \wave_{i,\waveidx}) 
 = \sum_{\waveidx=1}^{\numsampwave} \respfun_i(\wave_{i,\waveidx}) \ssif(\feaLRMSI, \ptHRHSI{}, \wave_{i,\waveidx})
\label{eq:ssif}
\end{align}
$\ssif$ can be decomposed into three neural implicit functions -- a pixel feature decoder $\decsif$, a spectral encoder $\encband$, and a spectral decoder $\decband$. The pixel feature decoder takes the 2D feature map of the input image $\feaLRMSI$ as well as one arbitrary pixel location $\ptHRHSI \in [-1,1] \cardprod [-1, 1]$ of $\imgHRHSI$ and maps them to a pixel hidden feature $\sifhid \in \Real^{\decdim}$ where $\decdim$ is the hidden pixel feature dimension (see Equation \ref{eq:sif}). Here, $\decsif$ can be any spatial implicit function such as  LIIF \cite{chen2021LIIF} and ITSRN \citep{yang2021ITSRN}.
\vspace{-0.1cm}
\begin{align}
\sifhid = \decsif(\feaLRMSI, \ptHRHSI{})
\label{eq:sif}
\end{align}
The spectral encoder $\encband(\wave_{i,\waveidx})$ encodes a wavelength $\wave_{i,\waveidx}$ sampled from any wavelength interval $\wavevecHRHSI_{i} = [\wave_{i,s}, \wave_{i,e}]  \in \wavematHRHSI$ into a spectral embedding $\embband{i,\waveidx} \in \Real^{\decdim}$. We can implement $\encband$ as any position encoder \citep{vaswani2017attention,mai2019space2vec}. Please refer to Appendix \ref{sec:posenc} for a detailed description.
\vspace{-0.1cm}
\begin{align}
\embband{i,\waveidx}  = \encband(\wave_{i,\waveidx})
\label{eq:specenc}
\end{align}
Finally, the spectral decoder $\decband(\embband{i,\waveidx};\sifhid)$ 
is a multilayer perceptron whose weights are modulated by the image feature embedding $\sifhid$. $\decband$ maps the spectral embedding $\embband{i,\waveidx}$ into a radiance value of $\wave_{i,\waveidx}$ at location $\ptHRHSI{}$, i.e., $\ptval_{\ptid,i,\waveidx} = \decband(\embband{i,\waveidx};\sifhid)$. So we have
\vspace{-0.1cm}
\begin{align}
\ptval_{\ptid,i} 
 = \sum_{\waveidx=1}^{\numsampwave} \respfun_i(\wave_{i,\waveidx}) \ssif(\feaLRMSI, \ptHRHSI{}, \wave_{i,\waveidx}) 
 = \sum_{\waveidx=1}^{\numsampwave} \respfun_i(\wave_{i,\waveidx}) \decband(\embband{i,\waveidx};\sifhid)
 = \sum_{\waveidx=1}^{\numsampwave} \respfun_i(\wave_{i,\waveidx}) \ptval_{\ptid,i,\waveidx}
\label{eq:ssifdecomp}
\end{align}
The response function $\respfun_i(\wave_{i,\waveidx})$ can be a learnable function or a predefined function based on the knowledge of the target HSI sensor. To make the learning easier, we pick a predefined function, e.g. a Gaussian distribution or a uniform distribution, for each band $\band_i$ by following \cite{chi2021spectral}.

\begin{figure*}[t!]
	\centering \tiny
	\vspace*{-0.2cm}
	\begin{subfigure}[b]{0.5\textwidth}  
		\centering 
		\includegraphics[width=\textwidth]{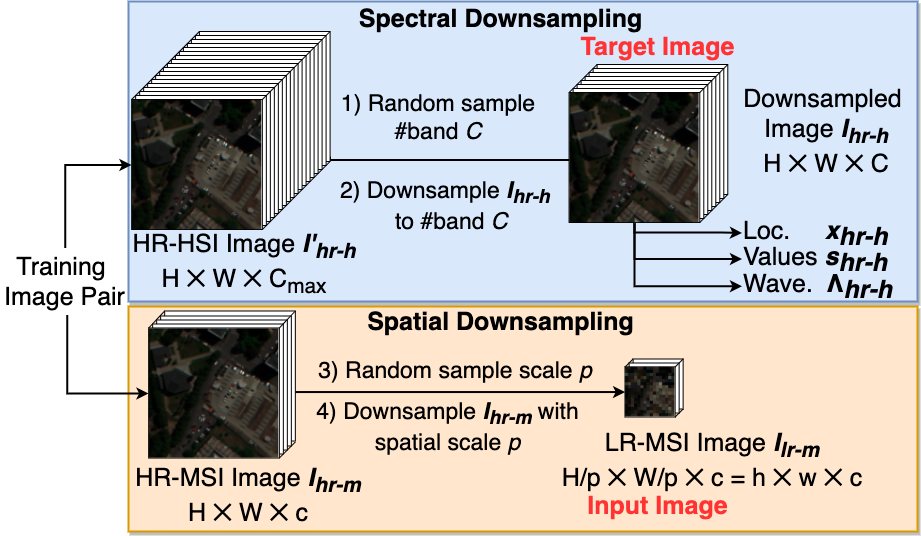}\vspace*{-0.2cm}
		\caption[]{{Data Preparation
		}}    
		\label{fig:model_data_prep}
	\end{subfigure}
\begin{subfigure}[b]{0.495\textwidth}  
		\centering 
		\includegraphics[width=\textwidth]{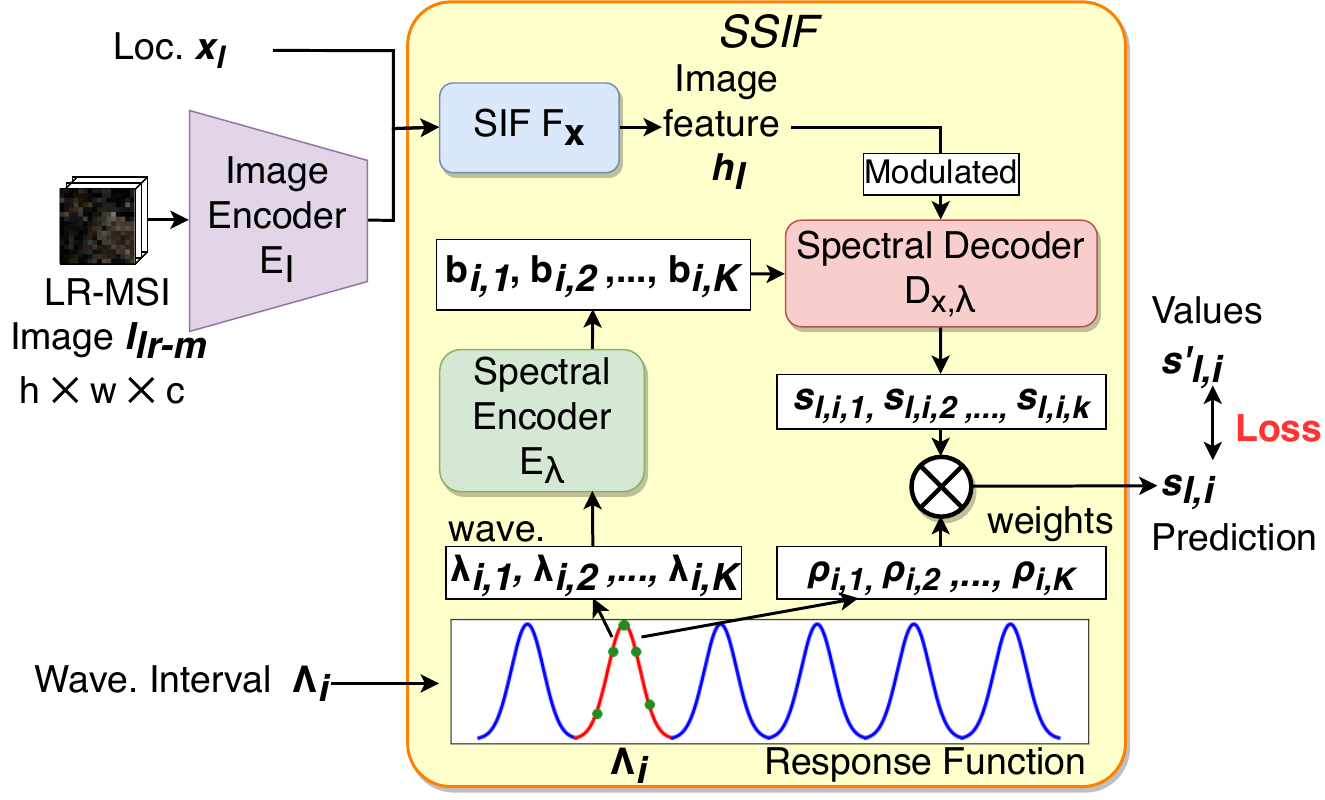}\vspace{-0.2cm}
		\caption[]{{$\modelname$ Training
		}}    
		\label{fig:model_arch}
	\end{subfigure}
	\vspace*{-0.3cm}
	\caption{
Data preparation (a) and training (b) for $\modelname$. In Figure (b), we use Gaussian distributions as the response functions for different wavelength intervals $\{\wavevecHRHSI_{1}, \wavevecHRHSI_{2},..,\wavevecHRHSI_{\chr}\}$ while the response function $\respfun_i(\wave_{i,\waveidx})$ for $\wavevecHRHSI_{i}$ is highlighted in red. 
    The green dots are $\numsampwave$ wavelengths $\{\wave_{i,1},\wave_{i,2},...,\wave_{i,\numsampwave}\}$ sampled from a wavelength interval $\wavevecHRHSI_{i} = [\wave_{i,s}, \wave_{i,e}]  \in \wavematHRHSI$ and $\{\respfun_{i,1},\respfun_{i,2},...,\respfun_{i,\numsampwave}\}$ are their corresponding response function values. $\{\embband{i,1},\embband{i,2},...,\embband{i,\numsampwave}\}$ are their encoded spectral embeddings. $\bigotimes$ indicates dot product as shown in Equation \ref{eq:ssifdecomp}.
}
	\label{fig:model}
    \vspace*{-0.3cm}
\end{figure*} 
Figure \ref{fig:model_arch} illustrates the model architecture of \modelname. The prediction $\ptval_{\ptid,i} \in \Real^{\chr}$ is compared with the ground truth $\ptvalgt{\ptid,i}$. 
A L1 reconstruction loss is used: \vspace{-0.2cm}
\begin{align}
\lossfun = \sum_{(\imgLRMSI, \imgHRHSI) \in \dataset} \sum_{(\ptHRHSI{}, \ptvalHRHSI{}, \wavematHRHSI) \in \imgHRHSI} \sum_{\wavevecHRHSI_{i} \in \wavematHRHSI} \parallel \ptval_{\ptid,i} - \ptvalgt{\ptid,i} \parallel_1,
\label{eq:l1loss}
\end{align}
where $\dataset$ indicates all the low-res and high-res image pairs for the 
SSSR task.

\subsection{Super-Resolution Data Preparation}  \label{sec:data_prep}

Figure \ref{fig:model_data_prep} illustrates the data preparation process of \modelname.
Given a training image pair which consists of a high spatial-spectral resolution image $\imgHRHSIMAX \in \Real^{\hhr \times \whr \times \cmax}$ and an image with high spatial resolution but low spectral resolution $\imgHRMSI \in \Real^{\hhr \times \whr \times \clr}$, we perform downsampling in both the spectral domain and spatial domain. For the spectral downsampling process (the blue box in Figure \ref{fig:model_data_prep}), we downsample $\imgHRHSIMAX$ in the spectral domain to obtain $\imgHRHSI \in \Real^{\hhr \times \whr \times \chr}$ where the band number $\chr$ is sampled uniformly between the min and max band number $\cmin, \cmax$.
For the spatial downsampling (the orange box in Figure \ref{fig:model_arch}), we spatially downsample $\imgHRMSI$ into $\imgLRMSI \in \Real^{\hlr \times \wlr \times \clr}$ which serves as the input for $\modelname$. Here, the downsampling scale $\scalespa$ is sampled uniformly from the min and max spatial scale $\scalespaMIN$, $ \scalespaMAX$. 
See Appendix \ref{sec:data_perp_app} for a detailed description.

\vspace{-0.3cm}
\section{Experiments}  \label{sec:exp}
\vspace{-0.3cm}
To test the effectiveness of the proposed \modelname, we evaluate it on two challenging spatial-spectral super-resolution benchmark datasets -- the CAVE dataset \citep{yasuma2010generalized} and the Pavia Centre dataset\footnote{\tiny{\url{http://www.ehu.eus/ccwintco/index.php/Hyperspectral_Remote_Sensing_Scenes}}}. Both datasets are widely used for super-resolution tasks on hyperspectral images. Please refer to Appendix \ref{sec:data} for detailed description of both datasets.
\vspace{-0.2cm}
\subsection{Baselines and \modelname{} Model Variants}  \label{sec:baseline}
\vspace{-0.2cm}
Compared with spatial SR and spectral SR, there has been much less work on spatiospectral super-resolution. So we mainly compare our model with 7 baselines: \textbf{RCAN + AWAN}, \textbf{AWAN + RCAN}, \textbf{AWAN + SSPSR}, \textbf{AWAN + SSPSR}, \textbf{RC/AW + MoG-DCN}, \textbf{RC/AW + MoG-DCN}, \textbf{SSJSR}, \textbf{US3RN}, and \textbf{LIIF}. Please refer to Appendix \ref{sec:baseline_app} for a detailed description for each baseline. 
For the first 6 baselines, we have to train separate SR models for different spatial and spectral resolutions of the output images. 
LIIF can use one model to generate output images with different spatial resolutions. However, we still need to train separate models when the output image $\imgHRMSI$ with different band numbers $\chr$. 
In contrast, our $\modelname$ model is able to handle different spatial and spectral resolutions with one model.

Based on the response functions we use (Gaussian or uniform) and the wavelength sampling methods, we have 4 \modelname{} variants: \textbf{\modelname{}-RF-GS}, \textbf{\modelname{}-RF-GF}, \textbf{\modelname{}-RF-US}, and \textbf{\modelname{}-RF-UF}. Both {\modelname{}-RF-GS} and {\modelname{}-RF-GF} uses a Gaussian distribution $\mathcal{N}(\mu_i,\,\sigma_i^{2})$ as the response function for each band $\band_i$ with wavelength interval $\wavevecHRHSI_{i} = [\wave_{i,s}, \wave_{i,e}]$ where $\mu_i = \frac{\wave_{i,s} + \wave_{i,e}}{2}$ and $\sigma_i = \frac{ \wave_{i,e} - \wave_{i,s}}{6}$. The difference is {\modelname{}-RF-GS} uses $\mathcal{N}(\mu_i,\,\sigma_i^{2})$ to sample $\numsampwave$ wavelengths from $\wavevecHRHSI_{i}$ while {\modelname{}-RF-GF} uses fixed $\numsampwave$ wavelengths with equal intervals in $\wavevecHRHSI_{i}$. Similarly, Both {\modelname{}-RF-US} and {\modelname{}-RF-UF} uses a Uniform distribution $\mathcal{U}(\wave_{i,s}, \wave_{i,e})$ as the response function for each band $\band_i$. {\modelname{}-RF-US} uses $\mathcal{U}(\wave_{i,s}, \wave_{i,e})$ to sample $\numsampwave$ wavelengths for each $\wavevecHRHSI_{i}$ while {\modelname{}-RF-UF} uses fixed $\numsampwave$ wavelengths with equal intervals.
We also consider 1 additional \modelname{} variant -- \textbf{\modelname{}-M} which only uses band middle point $\mu_i = \frac{\wave_{i,s} + \wave_{i,e}}{2}$ for each wavelength, i.e., $\numsampwave = 1$.

\vspace{-0.2cm}
\subsection{SSSR on the CAVE dataset} \label{sec:cave-res}
\vspace{-0.2cm}
Table \ref{tab:cave_eval_c31_short} shows the evaluation result of the SSSR task across different spatial scales $\scalespa$ on the original CAVE dataset with 31 bands. We use three evaluation metrics - PSNR, SSIM, and SAM which measure the quality of generated images from different perspectives. We evaluate different baselines as well as $\modelname$ under different spatial scales $\scalespa = \{2,4,8,10,12,14\}$. Since $\scalespaMIN = 1$ and $\scalespaMAX = 8$,  $\scalespa = \{2,4,8\}$ indicates "in-distribution" results while $\scalespa = \{10,12,14\}$ indicates "Out-of-distribution" results for $\scalespa$ not present to LIIF or $\modelname$ 
during training time. 
We can see that
\vspace{-0.5cm}
\begin{enumerate}[leftmargin=0cm]
\setlength\itemsep{-0.1em}
    \item All 5 $\modelname$ can outperform or are comparable to the 7 baselines across all tested spatial scales even if the first 6 baselines are trained separately on each $\scalespa$.
    
    \item {\modelname-RF-UF} achieves the best or 2nd best results across all spatial scales and metrics.

    \item A general pattern we can see across all spatial scales is that the order of the model performances is {\modelname-RF-*} $>$ {\modelname-M} $>$ {LIIF} $>$ other six baselines.

\end{enumerate}

\begin{figure*}[t!]
	\centering \tiny
	\vspace*{-0.2cm}
	\begin{subfigure}[b]{0.400\textwidth}  
		\centering 
		\includegraphics[width=\textwidth]{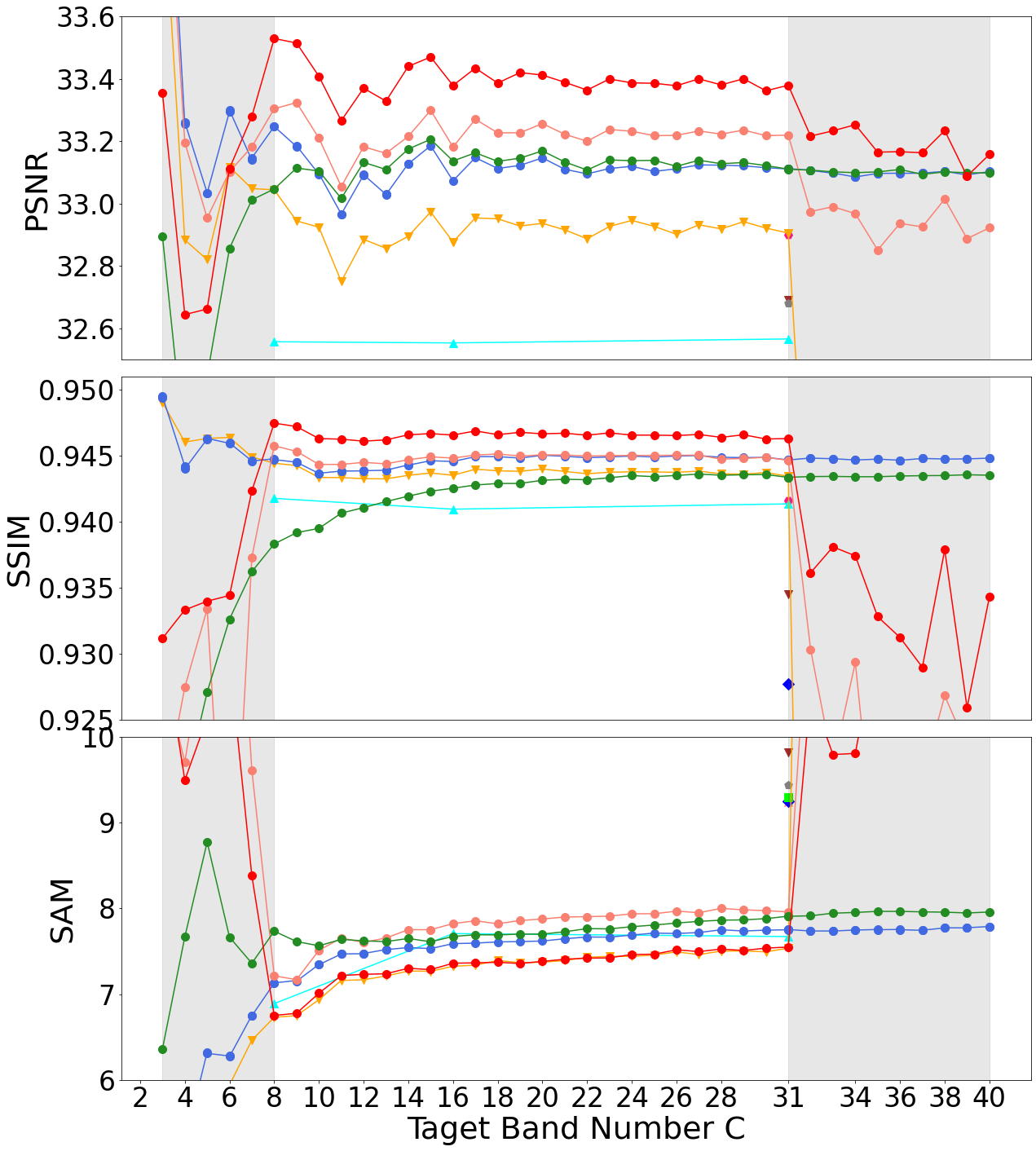}\vspace*{-0.2cm}
		\caption[]{{\small 
		Scale $\scalespa = 4$
		}}    
		\label{fig:cave_eval_scale4}
	\end{subfigure}
 	\hspace{0.05\textwidth}
	\begin{subfigure}[b]{0.535\textwidth}  
		\centering 
		\includegraphics[width=\textwidth]{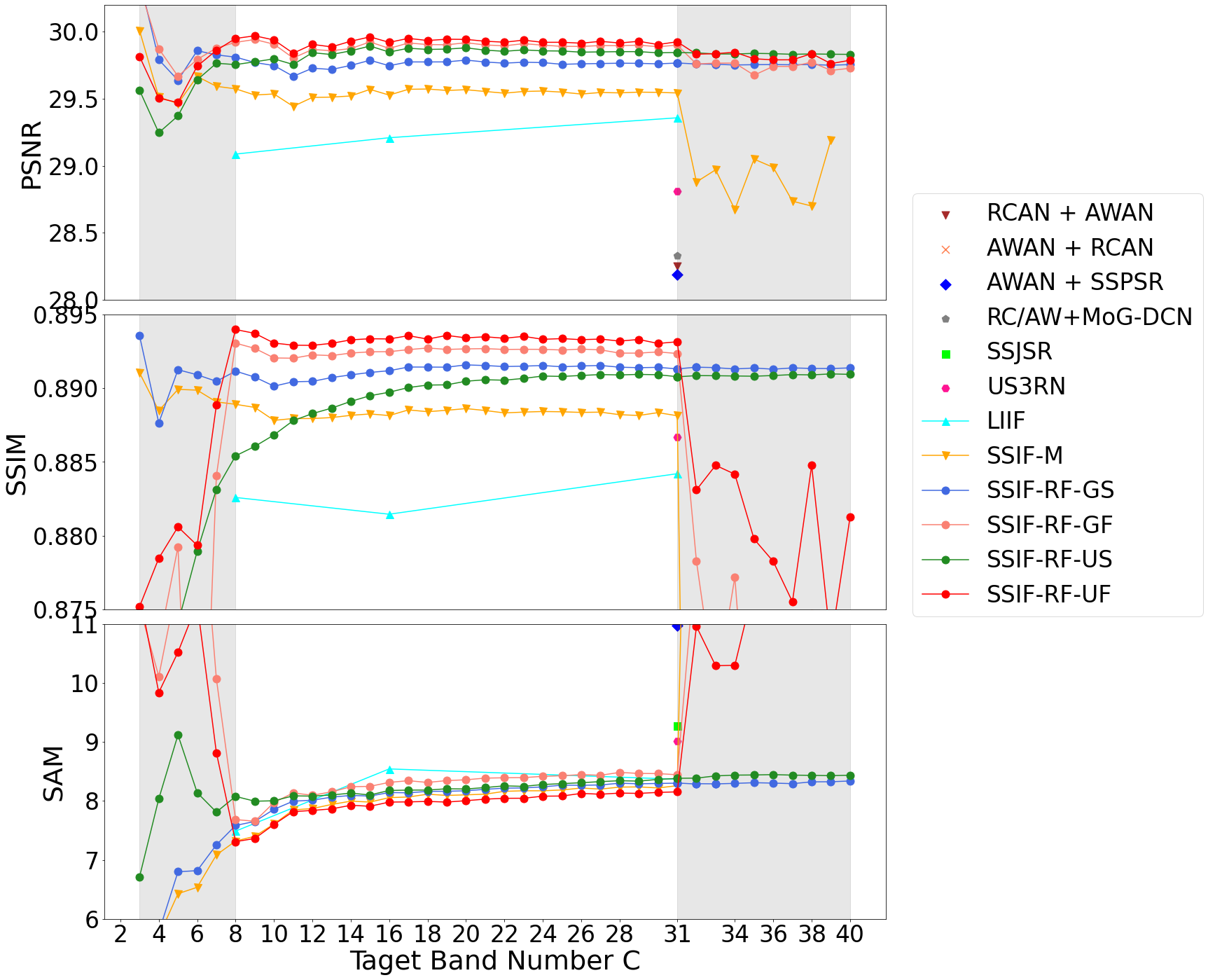}\vspace*{-0.2cm}
		\caption[]{{\small 
		Scale $\scalespa = 8$
		}}    
		\label{fig:cave_eval_scale8}
	\end{subfigure}
 \vspace*{-0.4cm}
	\caption{
	The evaluation result of the SSSR task across different $\chr$ on the CAVE \citep{yasuma2010CAVE} dataset. 
Here, the x axis indicates the number of bands $\chr$ of $\imgHRHSI$.
	(a) and (b) compare the performances of different models across different $\chr$ in two spatial scales $\scalespa=4$ or $\scalespa=8$.
	 Since our $\modelname$ can generalize to different $\scalespa$ and $\chr$, the evaluation metrics of each $\modelname$ are generated by one trained model. In contrast, we trained separated LIIF models for different $\chr$. The gray area in these plots indicates "out-of-distribution" performance in which $\modelname$ are evaluated on $\chr$s which have not been used for training.
}
	\label{fig:cave_eval_band}
    \vspace*{-0.4cm}
\end{figure*}

\begin{table}[]
\caption{
The evaluation result of the image super-resolution task across different spatial scales $\scalespa$ on the original CAVE \citep{yasuma2010CAVE} dataset with 31 bands. 
"In-distribution" and "Out-of-distribution" indicate whether the model has seen this spatial scale $\scalespa$ during training. This is only applicable to LIIF \citep{chen2021LIIF} and our different versions of $\modelname$ models. The performance of LIIF and \modelname~ across different $\scalespa$ are obtained from the same model while
for other 6 baselines, we trained separated SR models for each spatial scale $\scalespa$. Except for LIIF, the performances of all the other 6 baselines are from \citep{ma2021US3RN}.
We highlight the best model for each setting in bold and underline the second-best model. }
\vspace{-0.3cm}
	\label{tab:cave_eval_c31_short}
	\centering 
\tiny
	{\setlength{\tabcolsep}{3pt}
\begin{tabular}{l|ccc|ccc|ccc}
\toprule
Model          & \multicolumn{9}{c}{In-distribution}                                                                                                                    \\ \hline
Scale $\scalespa$        & \multicolumn{3}{c|}{2}                            & \multicolumn{3}{c|}{4}                            & \multicolumn{3}{c}{8}                            \\ \hline
Metric         & PSNR $\uparrow$           & SSIM $\uparrow$            & SAM $\downarrow$          & PSNR $\uparrow$           & SSIM $\uparrow$            & SAM $\downarrow$          & PSNR $\uparrow$           & SSIM $\uparrow$            & SAM $\downarrow$          \\ \hline
RCAN\citep{zhang2018RCAN} + AWAN\citep{li2020AWAN}   & 36.22 & 0.971          & 8.81          & 32.69          & 0.935          & 9.82          & 28.25          & 0.834          & 11.73         \\
AWAN\citep{li2020AWAN} + RCAN\citep{zhang2018RCAN}   & 36.09          & 0.969          & 8.42          & 31.44          & 0.916          & 9.24          & 27.77          & 0.837          & 12.39         \\
AWAN\citep{li2020AWAN} + SSPSR\citep{mei2020SepSSJSR}  & 36.16          & 0.969          & 8.49          & 32.34          & 0.928          & 9.25          & 28.19          & 0.860          & 10.97         \\
RC/AW+MoG-DCN\citep{dong2021MoGDCN} & 36.12          & 0.969          & 8.53          & 32.68          & 0.923          & 9.44          & 28.33          & 0.853           & 13.2          \\
SSJSR\citep{mei2020SepSSJSR}         & 35.51          & 0.970          & 7.67          & 30.9           & 0.916          & 9.3           & 27.3           & 0.844          & 9.28          \\
US3RN\citep{ma2021US3RN}         &  36.18    & { \ul 0.972}    & 7.43          & 32.9           & 0.942          &  7.91          & 28.81          &  0.887          & 9.02          \\
LIIF\citep{chen2021LIIF}          & 35.38          & 0.970            & {\ul 7.26}    & 32.57          & 0.941          & 7.67          & 29.36          & 0.884          & 8.37          \\ \hline
SSIF-M        & 35.80          & {\ul 0.972}    & 7.21          & 32.91          & 0.944          & \textbf{7.54} & 29.54          & 0.888          & {\ul 8.26}    \\
\hline
SSIF-RF-GS    & 36.29          & {\ul 0.972}    & 7.35          & 33.11          & {\ul 0.945}    & 7.75          & 29.77          & 0.891          & 8.30          \\
SSIF-RF-GF    & {\ul 36.37}    & {\ul 0.972}    & 7.49          & {\ul 33.22}    & {\ul 0.945}    & 7.96          & {\ul 29.90}    & {\ul 0.892}    & 8.45          \\
SSIF-RF-US    & 36.23          & 0.971          & 7.54          & 33.11          & 0.943          & 7.91          & 29.85          & 0.891          & 8.39          \\
SSIF-RF-UF    & \textbf{36.45} & \textbf{0.973} & \textbf{7.18} & \textbf{33.38} & \textbf{0.946} & {\ul 7.55}    & \textbf{29.93} & \textbf{0.893} & \textbf{8.16}     \\ \specialrule{.2em}{.1em}{.1em} 
Model         & \multicolumn{9}{c}{Out-of-distrobution}                                                                                                                \\ \hline
Scale $\scalespa$       & \multicolumn{3}{c|}{10}                           & \multicolumn{3}{c|}{12}                           & \multicolumn{3}{c|}{14}                           \\ \hline
Metric         & PSNR $\uparrow$           & SSIM $\uparrow$            & SAM $\downarrow$          & PSNR $\uparrow$           & SSIM $\uparrow$            & SAM $\downarrow$          & PSNR $\uparrow$           & SSIM $\uparrow$            & SAM $\downarrow$          \\ \hline
RCAN\citep{zhang2018RCAN} + AWAN\citep{li2020AWAN}   & -              & -               & -             & -              & -               & -             & -              & -               & -             \\
AWAN\citep{li2020AWAN} + RCAN\citep{zhang2018RCAN}   & -              & -               & -             & -              & -               & -             & -              & -               & -             \\
AWAN\citep{li2020AWAN} + SSPSR\citep{mei2020SepSSJSR}  & -              & -               & -             & -              & -               & -             & -              & -               & -             \\
RC/AW+MoG-DCN\citep{dong2021MoGDCN} & -              & -               & -             & -              & -               & -             & -              & -               & -             \\
SSJSR\citep{mei2020SepSSJSR}         & -              & -               & -             & -              & -               & -             & -              & -               & -             \\
US3RN\citep{ma2021US3RN}         & -              & -               & -             & -              & -               & -             & -              & -               & -             \\
LIIF\citep{chen2021LIIF}          & 27.59          & 0.859           &  8.62    & 26.67          & 0.838           & 8.96          &  25.5    & 0.822           & \textbf{9.17} \\ \hline
SSIF-M        & 27.94          & 0.865          & {\ul 8.54}    & 26.82          & 0.843          & 8.90          & 25.44          & 0.824          & 9.35          \\
\hline
SSIF-RF-GS    & 27.98          & 0.866          & 8.59          & 27.03          & {\ul 0.848}    & 8.95          & 25.50          & 0.828          & 9.45          \\
SSIF-RF-GF    & 28.05          & \textbf{0.869} & 8.56          & 26.96          & 0.847          & 8.94          & \textbf{25.67} & \textbf{0.830} & 9.34          \\
SSIF-RF-US    & \textbf{28.19} & {\ul 0.868}    & {\ul 8.54}    & {\ul 27.16}    & \textbf{0.849} & {\ul 8.88}    & 25.54          & {\ul 0.829}    & 9.36          \\
SSIF-RF-UF    & {\ul 28.14}    & \textbf{0.869} & \textbf{8.45} & \textbf{27.17} & \textbf{0.849} & \textbf{8.77} & {\ul 25.62}    & \textbf{0.830} & {\ul 9.19}                \\ \bottomrule  
\end{tabular}
}
	\vspace*{-0.3cm}
\end{table} 
\vspace{-0.3cm}
More interesting results emerge when we compare the performance of different models on different spectral resolutions, i.e., different $\chr$. Figure \ref{fig:cave_eval_scale4} and \ref{fig:cave_eval_scale8} compare model performance under different $\chr$ with a fixed spatial scale ($\scalespa = 4$ and $\scalespa = 8$ respectively). 
We can see that
\vspace{-0.4cm}
\begin{enumerate}[leftmargin=0cm]
\setlength\itemsep{-0.2em}
    \item Both Figure \ref{fig:cave_eval_scale4} and \ref{fig:cave_eval_scale8} show that \modelname-RF-UF achieves the best performances in two spatial scales and three metrics on "in-distribution" spectral resolutions. 
    \item However, the performance of \modelname-RF-UF, \modelname-RF-GF, and \modelname-M drop significantly when $\chr > 31$ while the performances of \modelname-RF-US and \modelname-RF-GS keep nearly unchanged for $\chr > 31$. This is because the first three \modelname{} use a fixed set of wavelengths during training while \modelname-RF-US and \modelname-RF-GS also sample novel wavelengths for each forward pass. This makes these two models have higher generalizability in "out-of-distribution" spectral scales. 
    \item A general pattern can be observed is that the order of model performance is \modelname{}-RF-* $>$ \modelname-M $>$ LIIF $>$ other six baselines.   
\end{enumerate}
Ablation studies on different designs of spectral decoder $\decband$ can be seen in Appendix \ref{sec:cave_res_app}.

\vspace{-0.2cm}
\subsection{SSSR on the Pavia Centre Remote Sensing dataset}  
\label{sec:ps_res}
\vspace{-0.2cm}

Table \ref{tab:pc_eval_c102_short} shows the evaluation results of the SSSR task across different spatial scales $\scalespa = \{2,3,4,8,10,12,14,16\}$ on the original Pavia Centre dataset with 102 bands. The setup is the same as Table \ref{tab:cave_eval_c31_short}. We can see that
\vspace{-0.4cm}
\begin{enumerate}[leftmargin=0cm]
\setlength\itemsep{0em}
    \item Except for $\scalespa = 2$, all \modelname{} can outperform all baselines on different spatial scales.
    \item  The performances of 4 \modelname-RF-* models are very similar across different spatial scales while \modelname-RF-US is the winner in most cases. They can outperform LIIF in most settings. \end{enumerate}

\begin{table}[]
\caption{
Image super-resolution
on the original Pavia Centre \citep{yasuma2010CAVE} dataset with 102 bands. We evaluate models across different spatial scales $\scalespa = \{2,3,4,8,10,12,14,16\}$. 
"In-distribution" and "Out-of-distribution" have the same meaning as Table \ref{tab:cave_eval_c31_short}.
The performance of LIIF and \modelname~ across different $\scalespa$ are obtained from the same model while other 6 baselines need to be trained separately or each $\scalespa$. 
Except for LIIF, the performances of all the other 6 baselines are from \citep{ma2021US3RN}.
$\modelname-M*$ and $\modelname-M$ treat each band as a point while other  $\modelname$ models treat each band as an interval. }
\vspace{-0.3cm}
	\label{tab:pc_eval_c102_short}
	\centering
	{\setlength{\tabcolsep}{1pt} 
\tiny
\begin{tabular}{l|ccc|ccc|ccc|ccc}
\toprule
Model              & \multicolumn{12}{c}{In-distribution}                                                                                                                                                                      \\ \hline
Scale $\scalespa$       & \multicolumn{3}{c|}{2}                            & \multicolumn{3}{c|}{3}                            & \multicolumn{3}{c|}{4}                            & \multicolumn{3}{c}{8}                            \\ \hline
Metric         & PSNR $\uparrow$           & SSIM $\uparrow$           & SAM $\downarrow$          & PSNR $\uparrow$          & SSIM $\uparrow$           & SAM $\downarrow$          & PSNR $\uparrow$           & SSIM $\uparrow$           & SAM $\downarrow$          & PSNR $\uparrow$          & SSIM $\uparrow$            & SAM $\downarrow$          \\ \hline
RCAN\citep{zhang2018RCAN} + AWAN\citep{li2020AWAN}    & 34.23          & 0.932          & 4.38          & 29.67          & 0.829          & 5.60          & 27.60          & 0.732          & 6.63       & 23.91          & 0.496          & 8.45          \\
AWAN\citep{li2020AWAN} + RCAN\citep{zhang2018RCAN}   & 34.54          & 0.936          & 4.38          & 29.66          & 0.827          & 5.70          & 27.61          & 0.734          & 6.69       & 23.67          & 0.515          & 8.87          \\
AWAN\citep{li2020AWAN} + SSPSR\citep{mei2020SepSSJSR}  & 34.24          & 0.934          & 4.30          & 29.60          & 0.828          & 5.55          & 27.71          & 0.742          & 6.32       & 24.21          & 0.506          & 8.14          \\
RC/AW+MoG-DCN\citep{dong2021MoGDCN} & 34.01          & 0.929          & 4.91          & 29.77          & 0.833          & 5.53          & 27.59          & 0.734          & 6.66       & 23.92          & 0.528          & 8.44          \\
SSJSR\citep{mei2020SepSSJSR}         & 31.80          & 0.894          & 4.80          & 29.05          & 0.810          & 6.14          & 27.06          & 0.703          & 6.93       & 20.61          & 0.347          & 18.30         \\
US3RN\citep{ma2021US3RN}         & \textbf{35.86} & 0.951          & 3.71          & 30.38          & 0.857          & 4.88          & 28.23          & 0.775          & 5.80       & 24.26          & 0.548          & 7.96          \\
LIIF\citep{chen2021LIIF}          & 35.24          & {\ul 0.952}    & 3.91          & 30.72          & 0.881          & 4.76          & 28.67          & 0.815          & 5.43       & 24.52          & 0.551          & 7.72          \\ \hline
SSIF-M        & 35.48          & \textbf{0.954} & 3.86          & \textbf{30.91} & \textbf{0.888} & {\ul 4.69}    & 28.76          & {\ul 0.820}    & 5.37       & 24.61          & {\ul 0.571}    & 7.63          \\
\hline
SSIF-RF-GS    & 35.47          & \textbf{0.954} & 3.87          & 30.87          & {\ul 0.887}    & 4.71          & \textbf{28.84} & \textbf{0.821} & 5.37       & 24.62          & 0.569          & 7.62          \\
SSIF-RF-GF    & {\ul 35.49}    & \textbf{0.954} & 3.85          & 30.85          & 0.886          & 4.70          & 28.81          & \textbf{0.821} & 5.35       & {\ul 24.64}    & \textbf{0.572} & {\ul 7.61}    \\
SSIF-RF-US    & 35.47          & \textbf{0.954} & 3.86          & \textbf{30.91} & 0.886          & {\ul 4.69}    & 28.81          & \textbf{0.821} & {\ul 5.36} & \textbf{24.66} & \textbf{0.572} & 7.62          \\
SSIF-RF-UF    & 35.48          & \textbf{0.954} & 3.84          & 30.88          & 0.886          & \textbf{4.68} & {\ul 28.83}    & \textbf{0.821} & 5.35       & 24.60          & 0.570          & \textbf{7.60} \\\specialrule{.2em}{.1em}{.1em} 
Model         & \multicolumn{12}{c}{Out-of-distribution}                                                                                                                                                                  \\ \hline
Scale $\scalespa$       & \multicolumn{3}{c|}{10}                           & \multicolumn{3}{c|}{12}                           & \multicolumn{3}{c|}{14}                           & \multicolumn{3}{c}{16}                           \\ \hline
Metric         & PSNR $\uparrow$           & SSIM $\uparrow$           & SAM $\downarrow$          & PSNR $\uparrow$          & SSIM $\uparrow$           & SAM $\downarrow$          & PSNR $\uparrow$           & SSIM $\uparrow$           & SAM $\downarrow$          & PSNR $\uparrow$          & SSIM $\uparrow$            & SAM $\downarrow$          \\ \hline
RCAN\citep{zhang2018RCAN} + AWAN\citep{li2020AWAN}   & -              & -               & -             & -              & -               & -             & -              & -               & -             & -              & -               & -             \\
AWAN\citep{li2020AWAN} + RCAN\citep{zhang2018RCAN}   & -              & -               & -             & -              & -               & -             & -              & -               & -             & -              & -               & -             \\
AWAN\citep{li2020AWAN} + SSPSR\citep{mei2020SepSSJSR}  & -              & -               & -             & -              & -               & -             & -              & -               & -             & -              & -               & -             \\
RC/AW+MoG-DCN\citep{dong2021MoGDCN} & -              & -               & -             & -              & -               & -             & -              & -               & -             & -              & -               & -             \\
SSJSR\citep{mei2020SepSSJSR}         & -              & -               & -             & -              & -               & -             & -              & -               & -             & -              & -               & -             \\
US3RN\citep{ma2021US3RN}         & -              & -               & -             & -              & -               & -             & -              & -               & -             & -              & -               & -             \\
LIIF\citep{chen2021LIIF}                    & 23.50          & 0.453          & 8.53          & 22.86          & 0.407          & 9.14          & 22.30          & 0.359          & \textbf{9.78}       & \textbf{22.10}          & {\ul 0.345}    & \textbf{9.91} \\ \hline
SSIF-M        & \textbf{23.53}          & {\ul 0.466}          & { \ul 8.47}          & 22.82          & 0.404          & 9.19          & { \ul 22.31}          & 0.359          & 9.85       & {\ul 22.05}    & 0.344          & 9.99          \\
\hline
SSIF-RF-GS    & 23.45          & 0.460          & 8.50          & 22.90          & 0.412          & 9.08          & 22.25          & 0.356          & 9.85       & 22.01          & 0.342          & 9.99          \\
SSIF-RF-GF    & \textbf{ 23.53}    & {\ul 0.466}          & \textbf{ 8.44}    & {\ul 22.96}    & { \ul 0.413}          & \textbf{9.01} & 22.22          & 0.350          & 9.86       & \textbf{22.10} & 0.340          & { \ul 9.94}          \\
SSIF-RF-US    & 23.46          & 0.458          & 8.57          & \textbf{22.99} & \textbf{0.415} & {\ul 9.03}    & \textbf{ 22.33}    & \textbf{0.363} & {\ul 9.82}       & \textbf{22.10} & \textbf{0.346} & 10.02         \\
SSIF-RF-UF    & {\ul 23.52}          & \textbf{ 0.468}    & 8.50          & 22.79          & 0.401          & 9.17          & 22.26          & {\ul 0.360}    & 9.85       & 22.01          & 0.343          & 10.04        \\
\bottomrule
\end{tabular}
}
	\vspace*{-0.3cm}
\end{table}

\begin{figure*}[t!]
	\centering \tiny
	\vspace*{-0.2cm}
	\begin{subfigure}[b]{0.400\textwidth}  
		\centering 
		\includegraphics[width=\textwidth]{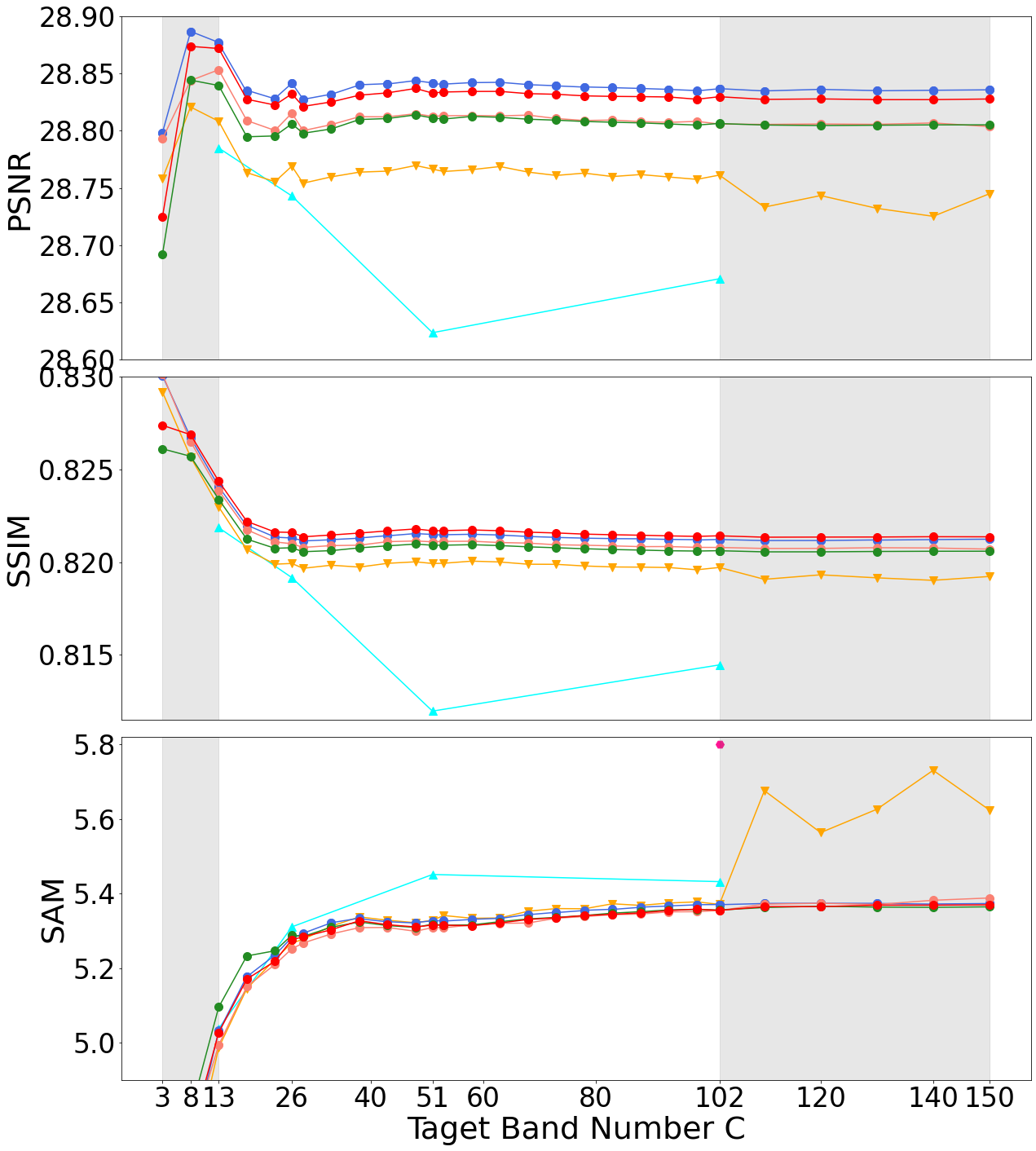}\vspace*{-0.2cm}
		\caption[]{{\small 
		Scale $\scalespa = 4$
		}}    
		\label{fig:pc_eval_scale4}
	\end{subfigure}
\hspace{0.05\textwidth}
	\begin{subfigure}[b]{0.535\textwidth}  
		\centering 
		\includegraphics[width=\textwidth]{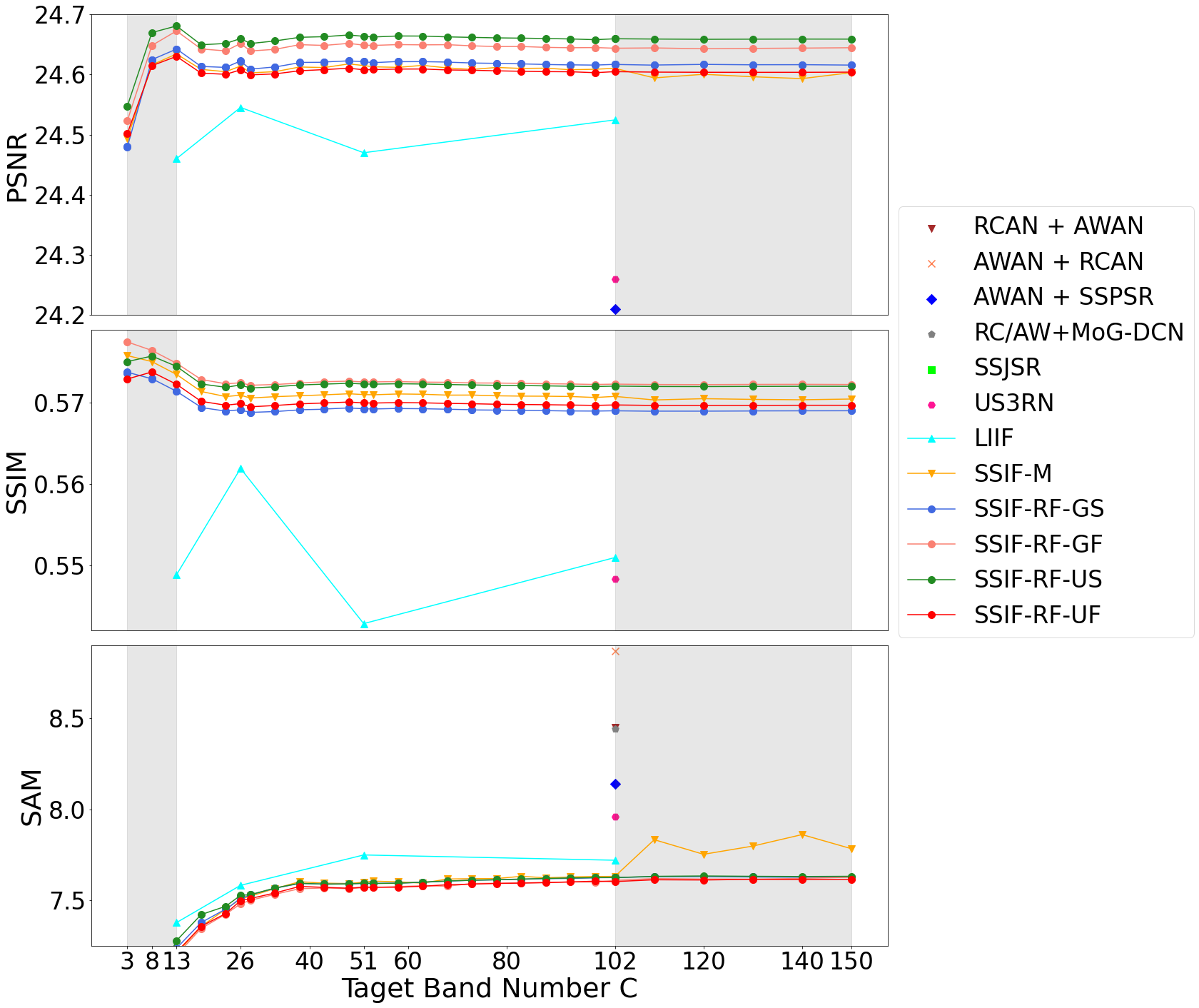}\vspace*{-0.2cm}
		\caption[]{{\small 
		Scale $\scalespa = 8$
		}}    
		\label{fig:pc_eval_scale8}
	\end{subfigure}
 \vspace*{-0.4cm}
	\caption{
	Evaluation across different $\chr$
on the Pavia Centre dataset. 
	The set-up is the same as Figure \ref{fig:cave_eval_band}.
Note that some of the baseline models do not appear in some of those plots because the performances of these models are very low and cannot be shown in the current metric range.
}
	\label{fig:pc_eval_band}
    \vspace*{-0.3cm}
\end{figure*}

Figure \ref{fig:pc_eval_scale4} and \ref{fig:pc_eval_scale8} compare different models across different spectral resolutions, i.e., $\chr$ for a fixed spatial scale ($\scalespa = 4$ and $\scalespa = 8$ respectively). We can see that
\begin{enumerate}[leftmargin=0cm]
\setlength\itemsep{-0.2em}
    \item The performances of 4 \modelname-RF-* models can outperform \modelname-M which is better than LIIF, and the other 6 baselines. 
    \item All 4 4 \modelname-RF-* show good generalizability for ``out-of-distribution'' spectral scales, especially when $\chr > 102$ while \modelname-M suffers from performance degradation.

\end{enumerate}

The ablation studies on $\numsampwave$ and the generated remote sensing images can be seen in Appendix \ref{sec:pc_res_app}. 

\vspace{-0.2cm}
\subsection{Land Use Classification on the Pavia Centre Dataset}  \label{sec:lulcc}
\vspace{-0.2cm}

To test the fidelity of the generated high spatial-spectral resolution images, we evaluate them on land use classification task.
We train the state-of-the-art land use classification model, A2S2K-ResNet \citep{roy2020A2S2K}, on the training dataset of Pavia Centre and evaluate its performance on the testing area -- both the ground truth HSI image as well as the generated images from LIIF and different \modelname~ models. Table \ref{tab:lulcc_c102_short} compares the performance of A2S2K-ResNet on different generated images across different spatial scales. 
We can see that although \modelname-M shows good performance on the SSSR task on both datasets, the generated images are less useful -- the land use classification accuracy on its generated images is much worse than other models, even far behind LIIF. \modelname-RF-GS shows the best performance across different spatial scales and can outperform LIIF by 1.7\%- 7\%.
Please refer to Appendix \ref{sec:lulcc-app} for a detailed description of the dataset, model, training detailed.

\begin{table}[]
\caption{
The evaluation of the generated images using A2S2K-ResNet \citep{roy2020A2S2K} on the Pavia Centre land use classification task.
"HSI" is the accuracy on the ground truth test image which is the upper bound.
"Acc Imp." is the accuracy improvement from LIIF to \modelname-RF-GS. 
}
\small
\vspace*{-0.4cm}
	\label{tab:lulcc_c102_short}
	\centering \small
\begin{tabular}{l|c|c|c|c}
\toprule
Model    & \multicolumn{4}{c}{Land Use Classification Accuracy (\%)} \\ \hline
Band $\chr $ & \multicolumn{4}{c}{102}                              \\ \hline
Scale $\scalespa$  & 2           & 3           & 4           & 8          \\ \hline
LIIF \citep{chen2021LIIF}              & 41.69                 & 41.29                 & 37.87                 & 37.38                 \\ \hline
SSIF-M            & 25.48                 & 25.38                 & 22.56                 & 14.91                 \\
\hline
SSIF-RF-GS        & {\ul 43.44}           & \textbf{46.86}        & \textbf{44.97}        & \textbf{44.82}        \\
SSIF-RF-GF        & 35.37                 & 37.91                 & 37.20                 & {\ul 38.08}           \\
SSIF-RF-US        & 40.15                 & 38.48                 & 34.86                 & 30.20                 \\
SSIF-RF-UF        & \textbf{45.32}        & {\ul 44.00}           & {\ul 41.87}           & 36.34                 \\ \hline
Acc Imp.          & 1.75                  & 5.57                  & 7.10                  & 7.44                  \\ \specialrule{.1em}{.1em}{.1em} 
HSI (Upper Bound)      & \multicolumn{4}{c}{72.66}                   \\ 
\bottomrule
\end{tabular}
	\vspace*{-0.5cm}
\end{table}

\vspace{-0.4cm}
\paragraph{Discussions of what the spectral encoder learned}
To understand how the spectral encoder represents a given wavelength we plot each dimension of spectral embedding against the wavelength (Figure \ref{fig:cave_specenc_dim} in Appendix \ref{sec:spec-basis}).  We find that they generally resemble piecewise-linear PL basis functions
\citep{1974pkb} or the continuous PK basis functions \citep{1976gpkb}. This makes sense because PL and PK are classical methods to represent a scalar function -- i.e., $\ssif(\feaLRMSI, \ptHRHSI{}, \param)$ in our case. We can think that the weights of these basis are provided by the image encoder and SIF network given an image $\feaLRMSI$ and location $\ptHRHSI{}$.
Having a spectral encoder with learnable parameters should provide better representation than fixed basis functions. \vspace{-0.3cm}
\section{Conclusion}  \label{sec:conclude}
\vspace{-0.3cm}
In this work, we propose \modelfullname~(\modelname), a neural implicit model that represents an image as a continuous function of both pixel coordinates in the spatial domain and wavelengths in the spectral domain. This enables \modelname~ to handle SSSR tasks with different output spatial and spectral resolutions simultaneously with one model. In contrast, all previous works have to train separate super-resolution models for different spectral resolutions.

We demonstrate the effectiveness of \modelname~ on the SSSR task with Two datasets -- CAVE and Pavia Centre. We show that \modelname~ can outperform all baselines across different spatial and spectral scales even when the baselines 
are allowed to be trained separately at each spectral resolution, thus solving an easier task.  
We demonstrate that \modelname{} generalizes well to unseen spatial and spectral resolutions. 
In addition, we test the fidelity of the generated images on a downstream task -- land use classification. We show that \modelname~ can outperform LIIF with a big margin (1.7-7\%).

In the current study, the effectiveness of \modelname~ is mainly shown on hyperspectral image SR, while \modelname~ is flexible enough to handle multispectral images with irregular wavelength intervals. This will be studied in future work. Moreover, the data limitation of the hyperspectral images poses a significant challenge to SR model training. We also plan to construct a large dataset for hyperspectral image super-resolution. %
 
\paragraph{Ethics Statement}
All datasets we use in this work including the CAVE and Pavia Centra datasets are publicly available datasets. No human subject study is conducted in this work. We do not find specific negative societal impacts of this work.

\paragraph{Reproducibility Statement}
Our source code has been uploaded as a supplementary file to reproduce our experimental results. The implementation details of the spectral encoder are described in Appendix \ref{sec:posenc}. The \modelname{} model training details are described in Appendix \ref{sec:training_detail_app}.


\begin{thebibliography}{73}
	\providecommand{\natexlab}[1]{#1}
	\providecommand{\url}[1]{\texttt{#1}}
	\expandafter\ifx\csname urlstyle\endcsname\relax
	\providecommand{\doi}[1]{doi: #1}\else
	\providecommand{\doi}{doi: \begingroup \urlstyle{rm}\Url}\fi
	
	\bibitem[Ayush et~al.(2021)Ayush, Uzkent, Meng, Tanmay, Burke, Lobell, and
	Ermon]{ayush2021geography}
	Kumar Ayush, Burak Uzkent, Chenlin Meng, Kumar Tanmay, Marshall Burke, David
	Lobell, and Stefano Ermon.
	\newblock Geography-aware self-supervised learning.
	\newblock In \emph{Proceedings of the IEEE/CVF International Conference on
		Computer Vision}, pp.\  10181--10190, 2021.
	
	\bibitem[Ball et~al.(2019)Ball, Chan, Christian, Jannuzi, Kim, Marrone,
	Medeiros, Ozel, Psaltis, Rose, et~al.]{2019ehtc}
	David Ball, Chi-kwan Chan, Pierre Christian, Buell~T Jannuzi, Junhan Kim,
	Daniel~P Marrone, Lia Medeiros, Feryal Ozel, Dimitrios Psaltis, Mel Rose,
	et~al.
	\newblock First m87 event horizon telescope results. i. the shadow of the
	supermassive black hole.
	\newblock \emph{Astrophysical Journal Letters}, 875\penalty0 (1), April 2019.
	\newblock ISSN 2041-8205.
	
	\bibitem[Barry et~al.(2002)Barry, Mendenhall, Jarecke, Folkman, Pearlman, and
	Markham]{barry2002eo1}
	Pamela~S Barry, J~Mendenhall, Peter Jarecke, Mark Folkman, J~Pearlman, and
	B~Markham.
	\newblock Eo-1 hyperion hyperspectral aggregation and comparison with eo-1
	advanced land imager and landsat 7 etm+.
	\newblock In \emph{IEEE International Geoscience and Remote Sensing Symposium},
	volume~3, pp.\  1648--1651. IEEE, 2002.
	
	\bibitem[Bioucas-Dias et~al.(2013)Bioucas-Dias, Plaza, Camps-Valls, Scheunders,
	Nasrabadi, and Chanussot]{Bioucas2013}
	Jose~M. Bioucas-Dias, Antonio Plaza, Gustavo Camps-Valls, Paul Scheunders,
	Nasser Nasrabadi, and Jocelyn Chanussot.
	\newblock Hyperspectral remote sensing data analysis and future challenges.
	\newblock \emph{IEEE Geoscience and Remote Sensing Magazine}, 1\penalty0
	(2):\penalty0 6--36, 2013.
	\newblock \doi{10.1109/MGRS.2013.2244672}.
	
	\bibitem[Brazile et~al.(2008)Brazile, Neville, Staenz, Schl{\"a}pfer, Sun, and
	Itten]{brazile2008toward}
	Jason Brazile, Robert~A Neville, Karl Staenz, Daniel Schl{\"a}pfer, Lixin Sun,
	and Klaus~I Itten.
	\newblock Toward scene-based retrieval of spectral response functions for
	hyperspectral imagers using fraunhofer features.
	\newblock \emph{Canadian Journal of Remote Sensing}, 34\penalty0
	(sup1):\penalty0 S43--S58, 2008.
	
	\bibitem[Chakraborty \& Trehan(2021)Chakraborty and
	Trehan]{chakraborty2021spectralnet}
	Tanmay Chakraborty and Utkarsh Trehan.
	\newblock Spectralnet: Exploring spatial-spectral waveletcnn for hyperspectral
	image classification.
	\newblock \emph{arXiv preprint arXiv:2104.00341}, 2021.
	
	\bibitem[Chen et~al.(2021)Chen, Liu, and Wang]{chen2021LIIF}
	Yinbo Chen, Sifei Liu, and Xiaolong Wang.
	\newblock Learning continuous image representation with local implicit image
	function.
	\newblock In \emph{Proceedings of the IEEE/CVF Conference on Computer Vision
		and Pattern Recognition}, pp.\  8628--8638, 2021.
	
	\bibitem[Chen \& Zhang(2019)Chen and Zhang]{chen2019learning}
	Zhiqin Chen and Hao Zhang.
	\newblock Learning implicit fields for generative shape modeling.
	\newblock In \emph{Proceedings of the IEEE/CVF Conference on Computer Vision
		and Pattern Recognition}, pp.\  5939--5948, 2019.
	
	\bibitem[Chi et~al.(2021)Chi, Lee, Hong, and Kim]{chi2021spectral}
	Junhwa Chi, Hyoungseok Lee, Soon~Gyu Hong, and Hyun-Cheol Kim.
	\newblock Spectral characteristics of the antarctic vegetation: A case study of
	barton peninsula.
	\newblock \emph{Remote Sensing}, 13\penalty0 (13):\penalty0 2470, 2021.
	
	\bibitem[Crawford et~al.(2019)Crawford, van~den Bosch, Brunt, Hom, Cooper,
	Harding, Butler, Dabney, Neumann, Cleckner, et~al.]{crawford2019radiometric}
	Christopher~J Crawford, Jeannette van~den Bosch, Kelly~M Brunt, Milton~G Hom,
	John~W Cooper, David~J Harding, James~J Butler, Philip~W Dabney, Thomas~A
	Neumann, Craig~S Cleckner, et~al.
	\newblock Radiometric calibration of a non-imaging airborne spectrometer to
	measure the greenland ice sheet surface.
	\newblock \emph{Atmospheric Measurement Techniques}, 12\penalty0 (3):\penalty0
	1913--1933, 2019.
	
	\bibitem[Cundill et~al.(2015)Cundill, van~der Werff, and Van~der
	Meijde]{cundill2015adjusting}
	Sharon~L Cundill, Harald~MA van~der Werff, and Mark Van~der Meijde.
	\newblock Adjusting spectral indices for spectral response function differences
	of very high spatial resolution sensors simulated from field spectra.
	\newblock \emph{Sensors}, 15\penalty0 (3):\penalty0 6221--6240, 2015.
	
	\bibitem[Deng et~al.(2021)Deng, Wang, Xu, Guo, Song, and Yang]{deng2021lau}
	Xin Deng, Hao Wang, Mai Xu, Yichen Guo, Yuhang Song, and Li~Yang.
	\newblock Lau-net: Latitude adaptive upscaling network for omnidirectional
	image super-resolution.
	\newblock In \emph{Proceedings of the IEEE/CVF Conference on Computer Vision
		and Pattern Recognition}, pp.\  9189--9198, 2021.
	
	\bibitem[Dong et~al.(2021)Dong, Zhou, Wu, Wu, Shi, and Li]{dong2021MoGDCN}
	Weisheng Dong, Chen Zhou, Fangfang Wu, Jinjian Wu, Guangming Shi, and Xin Li.
	\newblock Model-guided deep hyperspectral image super-resolution.
	\newblock \emph{IEEE Transactions on Image Processing}, 30:\penalty0
	5754--5768, 2021.
	
	\bibitem[Dupont et~al.(2021)Dupont, Goli{\'n}ski, Alizadeh, Teh, and
	Doucet]{dupont2021coin}
	Emilien Dupont, Adam Goli{\'n}ski, Milad Alizadeh, Yee~Whye Teh, and Arnaud
	Doucet.
	\newblock Coin: Compression with implicit neural representations.
	\newblock \emph{arXiv preprint arXiv:2103.03123}, 2021.
	
	\bibitem[Galliani et~al.(2017)Galliani, Lanaras, Marmanis, Baltsavias, and
	Schindler]{galliani2017learned}
	Silvano Galliani, Charis Lanaras, Dimitrios Marmanis, Emmanuel Baltsavias, and
	Konrad Schindler.
	\newblock Learned spectral super-resolution.
	\newblock \emph{arXiv preprint arXiv:1703.09470}, 2017.
	
	\bibitem[Han et~al.(2021)Han, Zhang, Xue, and Sun]{han2021spectral}
	Xiaolin Han, Huan Zhang, Jing-Hao Xue, and Weidong Sun.
	\newblock A spectral--spatial jointed spectral super-resolution and its
	application to hj-1a satellite images.
	\newblock \emph{IEEE Geoscience and Remote Sensing Letters}, 19:\penalty0 1--5,
	2021.
	
	\bibitem[Haris et~al.(2018)Haris, Shakhnarovich, and Ukita]{haris2018deep}
	Muhammad Haris, Gregory Shakhnarovich, and Norimichi Ukita.
	\newblock Deep back-projection networks for super-resolution.
	\newblock In \emph{Proceedings of the IEEE conference on computer vision and
		pattern recognition}, pp.\  1664--1673, 2018.
	
	\bibitem[He et~al.(2021)He, Wang, Lai, Zhang, Meng, Burke, Lobell, and
	Ermon]{he2021spatial}
	Yutong He, Dingjie Wang, Nicholas Lai, William Zhang, Chenlin Meng, Marshall
	Burke, David Lobell, and Stefano Ermon.
	\newblock Spatial-temporal super-resolution of satellite imagery via
	conditional pixel synthesis.
	\newblock \emph{Advances in Neural Information Processing Systems}, 34, 2021.
	
	\bibitem[Hu et~al.(2019)Hu, Mu, Zhang, Wang, Tan, and Sun]{hu2019MetaSR}
	Xuecai Hu, Haoyuan Mu, Xiangyu Zhang, Zilei Wang, Tieniu Tan, and Jian Sun.
	\newblock Meta-sr: A magnification-arbitrary network for super-resolution.
	\newblock In \emph{Proceedings of the IEEE/CVF Conference on Computer Vision
		and Pattern Recognition}, pp.\  1575--1584, 2019.
	
	\bibitem[Isaac \& Kulkarni(2015)Isaac and Kulkarni]{isaac2015super}
	Jithin~Saji Isaac and Ramesh Kulkarni.
	\newblock Super resolution techniques for medical image processing.
	\newblock In \emph{2015 International Conference on Technologies for
		Sustainable Development (ICTSD)}, pp.\  1--6. IEEE, 2015.
	
	\bibitem[Johnson et~al.(2007)Johnson, Wilson, Fink, Humayun, and
	Bearman]{Johnson2007SnapshotHI}
	William~R. Johnson, Daniel~W. Wilson, Wolfgang Fink, Mark~S. Humayun, and
	Gregory~H. Bearman.
	\newblock Snapshot hyperspectral imaging in ophthalmology.
	\newblock \emph{Journal of biomedical optics}, 12 1, 2007.
	
	\bibitem[Khrulkov \& Babenko(2021)Khrulkov and Babenko]{khrulkov2021neural}
	Valentin Khrulkov and Artem Babenko.
	\newblock Neural side-by-side: Predicting human preferences for no-reference
	super-resolution evaluation.
	\newblock In \emph{Proceedings of the IEEE/CVF Conference on Computer Vision
		and Pattern Recognition}, pp.\  4988--4997, 2021.
	
	\bibitem[Ledig et~al.(2017)Ledig, Theis, Husz{\'a}r, Caballero, Cunningham,
	Acosta, Aitken, Tejani, Totz, Wang, et~al.]{ledig2017photo}
	Christian Ledig, Lucas Theis, Ferenc Husz{\'a}r, Jose Caballero, Andrew
	Cunningham, Alejandro Acosta, Andrew Aitken, Alykhan Tejani, Johannes Totz,
	Zehan Wang, et~al.
	\newblock Photo-realistic single image super-resolution using a generative
	adversarial network.
	\newblock In \emph{Proceedings of the IEEE conference on computer vision and
		pattern recognition}, pp.\  4681--4690, 2017.
	
	\bibitem[Li et~al.(2020)Li, Wu, Song, Li, and Liu]{li2020AWAN}
	Jiaojiao Li, Chaoxiong Wu, Rui Song, Yunsong Li, and Fei Liu.
	\newblock Adaptive weighted attention network with camera spectral sensitivity
	prior for spectral reconstruction from rgb images.
	\newblock In \emph{Proceedings of the IEEE/CVF Conference on Computer Vision
		and Pattern Recognition Workshops}, pp.\  462--463, 2020.
	
	\bibitem[Lim et~al.(2017)Lim, Son, Kim, Nah, and Mu~Lee]{lim2017EDSR}
	Bee Lim, Sanghyun Son, Heewon Kim, Seungjun Nah, and Kyoung Mu~Lee.
	\newblock Enhanced deep residual networks for single image super-resolution.
	\newblock In \emph{Proceedings of the IEEE conference on computer vision and
		pattern recognition workshops}, pp.\  136--144, 2017.
	
	\bibitem[Lu \& Fei(2014)Lu and Fei]{Lu2014MedicalHI}
	Guolan Lu and Baowei Fei.
	\newblock Medical hyperspectral imaging: a review.
	\newblock \emph{Journal of Biomedical Optics}, 19, 2014.
	
	\bibitem[Ma et~al.(2021)Ma, Jiang, Liu, and Ma]{ma2021US3RN}
	Qing Ma, Junjun Jiang, Xianming Liu, and Jiayi Ma.
	\newblock Deep unfolding network for spatiospectral image super-resolution.
	\newblock \emph{IEEE Transactions on Computational Imaging}, 2021.
	
	\bibitem[Mai et~al.(2020{\natexlab{a}})Mai, Janowicz, Cai, Zhu, Regalia, Yan,
	Shi, and Lao]{mai2020se}
	Gengchen Mai, Krzysztof Janowicz, Ling Cai, Rui Zhu, Blake Regalia, Bo~Yan,
	Meilin Shi, and Ni~Lao.
	\newblock Se-kge: A location-aware knowledge graph embedding model for
	geographic question answering and spatial semantic lifting.
	\newblock \emph{Transactions in GIS}, 24\penalty0 (3):\penalty0 623--655,
	2020{\natexlab{a}}.
	
	\bibitem[Mai et~al.(2020{\natexlab{b}})Mai, Janowicz, Yan, Zhu, Cai, and
	Lao]{mai2019space2vec}
	Gengchen Mai, Krzysztof Janowicz, Bo~Yan, Rui Zhu, Ling Cai, and Ni~Lao.
	\newblock Multi-scale representation learning for spatial feature distributions
	using grid cells.
	\newblock In \emph{International Conference on Learning Representations},
	2020{\natexlab{b}}.
	
	\bibitem[Mai et~al.(2022)Mai, Janowicz, Hu, Gao, Yan, Zhu, Cai, and
	Lao]{mai2022review}
	Gengchen Mai, Krzysztof Janowicz, Yingjie Hu, Song Gao, Bo~Yan, Rui Zhu, Ling
	Cai, and Ni~Lao.
	\newblock A review of location encoding for geoai: methods and applications.
	\newblock \emph{International Journal of Geographical Information Science},
	pp.\  1--35, 2022.
	
	\bibitem[Mai et~al.(2023{\natexlab{a}})Mai, Lao, He, Song, and
	Ermon]{mai2023csp}
	Gengchen Mai, Ni~Lao, Yutong He, Jiaming Song, and Stefano Ermon.
	\newblock Csp: Self-supervised contrastive spatial pre-training for
	geospatial-visual representations.
	\newblock In \emph{the Fortieth International Conference on Machine Learning
		(ICML 2023)}, 2023{\natexlab{a}}.
	
	\bibitem[Mai et~al.(2023{\natexlab{b}})Mai, Xuan, Zuo, He, Song, Ermon,
	Janowicz, and Lao]{mai2023sphere2vec}
	Gengchen Mai, Yao Xuan, Wenyun Zuo, Yutong He, Jiaming Song, Stefano Ermon,
	Krzysztof Janowicz, and Ni~Lao.
	\newblock Sphere2vec: A general-purpose location representation learning over a
	spherical surface for large-scale geospatial predictions.
	\newblock \emph{ISPRS Journal of Photogrammetry and Remote Sensing},
	202:\penalty0 439--462, 2023{\natexlab{b}}.
	
	\bibitem[Mei et~al.(2017)Mei, Yuan, Ji, Zhang, Wan, and
	Du]{mei2017hyperspectral}
	Shaohui Mei, Xin Yuan, Jingyu Ji, Yifan Zhang, Shuai Wan, and Qian Du.
	\newblock Hyperspectral image spatial super-resolution via 3d full
	convolutional neural network.
	\newblock \emph{Remote Sensing}, 9\penalty0 (11):\penalty0 1139, 2017.
	
	\bibitem[Mei et~al.(2020)Mei, Jiang, Li, and Du]{mei2020SepSSJSR}
	Shaohui Mei, Ruituo Jiang, Xu~Li, and Qian Du.
	\newblock Spatial and spectral joint super-resolution using convolutional
	neural network.
	\newblock \emph{IEEE Transactions on Geoscience and Remote Sensing},
	58\penalty0 (7):\penalty0 4590--4603, 2020.
	
	\bibitem[Melal(1976)]{1976gpkb}
	C.~Melal.
	\newblock Generalized paul-koch basis functions.
	\newblock \emph{IEEE Transactions on Acoustics, Speech, and Signal Processing},
	24\penalty0 (3):\penalty0 263--264, 1976.
	\newblock \doi{10.1109/TASSP.1976.1162804}.
	
	\bibitem[Melgani \& Bruzzone(2004)Melgani and Bruzzone]{Melgani2004}
	F.~Melgani and L.~Bruzzone.
	\newblock Classification of hyperspectral remote sensing images with support
	vector machines.
	\newblock \emph{IEEE Transactions on Geoscience and Remote Sensing},
	42\penalty0 (8):\penalty0 1778--1790, 2004.
	\newblock \doi{10.1109/TGRS.2004.831865}.
	
	\bibitem[Mescheder et~al.(2019)Mescheder, Oechsle, Niemeyer, Nowozin, and
	Geiger]{mescheder2019occupancy}
	Lars Mescheder, Michael Oechsle, Michael Niemeyer, Sebastian Nowozin, and
	Andreas Geiger.
	\newblock Occupancy networks: Learning 3d reconstruction in function space.
	\newblock In \emph{Proceedings of the IEEE/CVF Conference on Computer Vision
		and Pattern Recognition}, pp.\  4460--4470, 2019.
	
	\bibitem[Mildenhall et~al.(2020)Mildenhall, Srinivasan, Tancik, Barron,
	Ramamoorthi, and Ng]{mildenhall2020nerf}
	Ben Mildenhall, Pratul~P Srinivasan, Matthew Tancik, Jonathan~T Barron, Ravi
	Ramamoorthi, and Ren Ng.
	\newblock Nerf: Representing scenes as neural radiance fields for view
	synthesis.
	\newblock In \emph{European conference on computer vision}, pp.\  405--421.
	Springer, 2020.
	
	\bibitem[Niu et~al.(2020)Niu, Wen, Ren, Zhang, Yang, Wang, Zhang, Cao, and
	Shen]{niu2020single}
	Ben Niu, Weilei Wen, Wenqi Ren, Xiangde Zhang, Lianping Yang, Shuzhen Wang,
	Kaihao Zhang, Xiaochun Cao, and Haifeng Shen.
	\newblock Single image super-resolution via a holistic attention network.
	\newblock In \emph{European conference on computer vision}, pp.\  191--207.
	Springer, 2020.
	
	\bibitem[Pan et~al.(2003)Pan, Healey, Prasad, and Tromberg]{2003pan}
	Zhihong Pan, G.~Healey, M.~Prasad, and B.~Tromberg.
	\newblock Face recognition in hyperspectral images.
	\newblock \emph{IEEE Transactions on Pattern Analysis and Machine
		Intelligence}, 25\penalty0 (12):\penalty0 1552--1560, 2003.
	\newblock \doi{10.1109/TPAMI.2003.1251148}.
	
	\bibitem[Park et~al.(2019)Park, Florence, Straub, Newcombe, and
	Lovegrove]{park2019deepsdf}
	Jeong~Joon Park, Peter Florence, Julian Straub, Richard Newcombe, and Steven
	Lovegrove.
	\newblock Deepsdf: Learning continuous signed distance functions for shape
	representation.
	\newblock In \emph{Proceedings of the IEEE/CVF Conference on Computer Vision
		and Pattern Recognition}, pp.\  165--174, 2019.
	
	\bibitem[Paul \& Koch(1974)Paul and Koch]{1974pkb}
	C.~Paul and R.~Koch.
	\newblock On piecewise-linear basis functions and piecewise-linear signal
	expansions.
	\newblock \emph{IEEE Transactions on Acoustics, Speech, and Signal Processing},
	22\penalty0 (4):\penalty0 263--268, 1974.
	\newblock \doi{10.1109/TASSP.1974.1162585}.
	
	\bibitem[Qu et~al.(2021)Qu, Qi, Kwan, Yokoya, and
	Chanussot]{qu2021unsupervised}
	Ying Qu, Hairong Qi, Chiman Kwan, Naoto Yokoya, and Jocelyn Chanussot.
	\newblock Unsupervised and unregistered hyperspectral image super-resolution
	with mutual dirichlet-net.
	\newblock \emph{IEEE Transactions on Geoscience and Remote Sensing},
	60:\penalty0 1--18, 2021.
	
	\bibitem[Roy et~al.(2020)Roy, Manna, Song, and Bruzzone]{roy2020A2S2K}
	Swalpa~Kumar Roy, Suvojit Manna, Tiecheng Song, and Lorenzo Bruzzone.
	\newblock Attention-based adaptive spectral--spatial kernel resnet for
	hyperspectral image classification.
	\newblock \emph{IEEE Transactions on Geoscience and Remote Sensing},
	59\penalty0 (9):\penalty0 7831--7843, 2020.
	
	\bibitem[Saharia et~al.(2021)Saharia, Ho, Chan, Salimans, Fleet, and
	Norouzi]{saharia2021SR3}
	Chitwan Saharia, Jonathan Ho, William Chan, Tim Salimans, David~J Fleet, and
	Mohammad Norouzi.
	\newblock Image super-resolution via iterative refinement.
	\newblock \emph{arXiv preprint arXiv:2104.07636}, 2021.
	
	\bibitem[Sitzmann et~al.(2020)Sitzmann, Martel, Bergman, Lindell, and
	Wetzstein]{sitzmann2020implicit}
	Vincent Sitzmann, Julien Martel, Alexander Bergman, David Lindell, and Gordon
	Wetzstein.
	\newblock Implicit neural representations with periodic activation functions.
	\newblock \emph{Advances in Neural Information Processing Systems},
	33:\penalty0 7462--7473, 2020.
	
	\bibitem[Strudel et~al.(2021)Strudel, Garcia, Laptev, and
	Schmid]{strudel2021segmenter}
	Robin Strudel, Ricardo Garcia, Ivan Laptev, and Cordelia Schmid.
	\newblock Segmenter: Transformer for semantic segmentation.
	\newblock In \emph{Proceedings of the IEEE/CVF International Conference on
		Computer Vision}, pp.\  7262--7272, 2021.
	
	\bibitem[Str{\"u}mpler et~al.(2021)Str{\"u}mpler, Postels, Yang, Van~Gool, and
	Tombari]{strumpler2021implicit}
	Yannick Str{\"u}mpler, Janis Postels, Ren Yang, Luc Van~Gool, and Federico
	Tombari.
	\newblock Implicit neural representations for image compression.
	\newblock \emph{arXiv preprint arXiv:2112.04267}, 2021.
	
	\bibitem[Tancik et~al.(2020)Tancik, Srinivasan, Mildenhall, Fridovich-Keil,
	Raghavan, Singhal, Ramamoorthi, Barron, and Ng]{tancik2020fourier}
	Matthew Tancik, Pratul Srinivasan, Ben Mildenhall, Sara Fridovich-Keil, Nithin
	Raghavan, Utkarsh Singhal, Ravi Ramamoorthi, Jonathan Barron, and Ren Ng.
	\newblock Fourier features let networks learn high frequency functions in low
	dimensional domains.
	\newblock \emph{Advances in Neural Information Processing Systems},
	33:\penalty0 7537--7547, 2020.
	
	\bibitem[Uzair et~al.(2015)Uzair, Mahmood, and Mian]{2015uzair}
	Muhammad Uzair, Arif Mahmood, and Ajmal Mian.
	\newblock Hyperspectral face recognition with spatiospectral information fusion
	and pls regression.
	\newblock \emph{IEEE Transactions on Image Processing}, 24\penalty0
	(3):\penalty0 1127--1137, 2015.
	\newblock \doi{10.1109/TIP.2015.2393057}.
	
	\bibitem[Van~Etten et~al.(2018)Van~Etten, Lindenbaum, and
	Bacastow]{van2018spacenet}
	Adam Van~Etten, Dave Lindenbaum, and Todd~M Bacastow.
	\newblock Spacenet: A remote sensing dataset and challenge series.
	\newblock \emph{arXiv preprint arXiv:1807.01232}, 2018.
	
	\bibitem[Vaswani et~al.(2017)Vaswani, Shazeer, Parmar, Uszkoreit, Jones, Gomez,
	Kaiser, and Polosukhin]{vaswani2017attention}
	Ashish Vaswani, Noam Shazeer, Niki Parmar, Jakob Uszkoreit, Llion Jones,
	Aidan~N Gomez, {\L}ukasz Kaiser, and Illia Polosukhin.
	\newblock Attention is all you need.
	\newblock \emph{Advances in neural information processing systems}, 30, 2017.
	
	\bibitem[Wang et~al.(2022{\natexlab{a}})Wang, Ma, and Jiang]{Wang2022}
	Xinya Wang, Jiayi Ma, and Junjun Jiang.
	\newblock Hyperspectral image super-resolution via recurrent feedback embedding
	and spatial–spectral consistency regularization.
	\newblock \emph{IEEE Transactions on Geoscience and Remote Sensing},
	60:\penalty0 1--13, 2022{\natexlab{a}}.
	\newblock \doi{10.1109/TGRS.2021.3064450}.
	
	\bibitem[Wang et~al.(2022{\natexlab{b}})Wang, Chen, and
	Richard]{wang2022hyperspectral}
	Xiuheng Wang, Jie Chen, and C{\'e}dric Richard.
	\newblock Hyperspectral image super-resolution with deep priors and degradation
	model inversion.
	\newblock \emph{arXiv preprint arXiv:2201.09851}, 2022{\natexlab{b}}.
	
	\bibitem[Wu et~al.(2021{\natexlab{a}})Wu, Ni, and Zhang]{wu2021SADN}
	Hanlin Wu, Ning Ni, and Libao Zhang.
	\newblock Scale-aware dynamic network for continuous-scale super-resolution.
	\newblock \emph{arXiv preprint arXiv:2110.15655}, 2021{\natexlab{a}}.
	
	\bibitem[Wu et~al.(2021{\natexlab{b}})Wu, Ni, and Zhang]{wu2021scale}
	Hanlin Wu, Ning Ni, and Libao Zhang.
	\newblock Scale-aware dynamic network for continuous-scale super-resolution.
	\newblock \emph{arXiv preprint arXiv:2110.15655}, 2021{\natexlab{b}}.
	
	\bibitem[Xiong et~al.(2020)Xiong, Zhou, and Qian]{2020xiong}
	Fengchao Xiong, Jun Zhou, and Yuntao Qian.
	\newblock Material based object tracking in hyperspectral videos.
	\newblock \emph{IEEE Transactions on Image Processing}, 29:\penalty0
	3719--3733, 2020.
	\newblock \doi{10.1109/TIP.2020.2965302}.
	
	\bibitem[Yang et~al.(2021)Yang, Shen, Yue, and Li]{yang2021ITSRN}
	Jingyu Yang, Sheng Shen, Huanjing Yue, and Kun Li.
	\newblock Implicit transformer network for screen content image continuous
	super-resolution.
	\newblock \emph{Advances in Neural Information Processing Systems}, 34, 2021.
	
	\bibitem[Yao et~al.(2020)Yao, Hong, Chanussot, Meng, Zhu, and Xu]{yao2020cross}
	Jing Yao, Danfeng Hong, Jocelyn Chanussot, Deyu Meng, Xiaoxiang Zhu, and
	Zongben Xu.
	\newblock Cross-attention in coupled unmixing nets for unsupervised
	hyperspectral super-resolution.
	\newblock In \emph{European Conference on Computer Vision}, pp.\  208--224.
	Springer, 2020.
	
	\bibitem[Yasuma et~al.(2010{\natexlab{a}})Yasuma, Mitsunaga, Iso, and
	Nayar]{yasuma2010CAVE}
	Fumihito Yasuma, Tomoo Mitsunaga, Daisuke Iso, and Shree~K Nayar.
	\newblock Generalized assorted pixel camera: postcapture control of resolution,
	dynamic range, and spectrum.
	\newblock \emph{IEEE transactions on image processing}, 19\penalty0
	(9):\penalty0 2241--2253, 2010{\natexlab{a}}.
	
	\bibitem[Yasuma et~al.(2010{\natexlab{b}})Yasuma, Mitsunaga, Iso, and
	Nayar]{yasuma2010generalized}
	Fumihito Yasuma, Tomoo Mitsunaga, Daisuke Iso, and Shree~K Nayar.
	\newblock Generalized assorted pixel camera: postcapture control of resolution,
	dynamic range, and spectrum.
	\newblock \emph{IEEE transactions on image processing}, 19\penalty0
	(9):\penalty0 2241--2253, 2010{\natexlab{b}}.
	
	\bibitem[Yoon et~al.(2021)Yoon, Chung, Wang, and Yoon]{yoon2021spheresr}
	Youngho Yoon, Inchul Chung, Lin Wang, and Kuk-Jin Yoon.
	\newblock Spheresr: 360◦ image super-resolution with arbitrary projection via
	continuous spherical image representation.
	\newblock \emph{arXiv e-prints}, pp.\  arXiv--2112, 2021.
	
	\bibitem[Zhang et~al.(2020{\natexlab{a}})Zhang, Gool, and
	Timofte]{zhang2020USRNet}
	Kai Zhang, Luc~Van Gool, and Radu Timofte.
	\newblock Deep unfolding network for image super-resolution.
	\newblock In \emph{Proceedings of the IEEE/CVF conference on computer vision
		and pattern recognition}, pp.\  3217--3226, 2020{\natexlab{a}}.
	
	\bibitem[Zhang et~al.(2020{\natexlab{b}})Zhang, Gool, and
	Timofte]{zhang2020deep}
	Kai Zhang, Luc~Van Gool, and Radu Timofte.
	\newblock Deep unfolding network for image super-resolution.
	\newblock In \emph{Proceedings of the IEEE/CVF conference on computer vision
		and pattern recognition}, pp.\  3217--3226, 2020{\natexlab{b}}.
	
	\bibitem[Zhang(2021)]{zhang2021implicit}
	Kaiwei Zhang.
	\newblock Implicit neural representation learning for hyperspectral image
	super-resolution.
	\newblock \emph{arXiv preprint arXiv:2112.10541}, 2021.
	
	\bibitem[Zhang et~al.(2020{\natexlab{c}})Zhang, Nie, Wei, Zhang, Liao, and
	Shao]{zhang2020unsupervised}
	Lei Zhang, Jiangtao Nie, Wei Wei, Yanning Zhang, Shengcai Liao, and Ling Shao.
	\newblock Unsupervised adaptation learning for hyperspectral imagery
	super-resolution.
	\newblock In \emph{Proceedings of the IEEE/CVF Conference on Computer Vision
		and Pattern Recognition}, pp.\  3073--3082, 2020{\natexlab{c}}.
	
	\bibitem[Zhang et~al.(2018{\natexlab{a}})Zhang, Li, Li, Wang, Zhong, and
	Fu]{zhang2018RCAN}
	Yulun Zhang, Kunpeng Li, Kai Li, Lichen Wang, Bineng Zhong, and Yun Fu.
	\newblock Image super-resolution using very deep residual channel attention
	networks.
	\newblock In \emph{Proceedings of the European conference on computer vision
		(ECCV)}, pp.\  286--301, 2018{\natexlab{a}}.
	
	\bibitem[Zhang et~al.(2018{\natexlab{b}})Zhang, Tian, Kong, Zhong, and
	Fu]{zhang2018RDN}
	Yulun Zhang, Yapeng Tian, Yu~Kong, Bineng Zhong, and Yun Fu.
	\newblock Residual dense network for image super-resolution.
	\newblock In \emph{Proceedings of the IEEE conference on computer vision and
		pattern recognition}, pp.\  2472--2481, 2018{\natexlab{b}}.
	
	\bibitem[Zhao et~al.(2018)Zhao, Zhang, Liu, Shi, Loy, Lin, and
	Jia]{zhao2018psanet}
	Hengshuang Zhao, Yi~Zhang, Shu Liu, Jianping Shi, Chen~Change Loy, Dahua Lin,
	and Jiaya Jia.
	\newblock Psanet: Point-wise spatial attention network for scene parsing.
	\newblock In \emph{Proceedings of the European conference on computer vision
		(ECCV)}, pp.\  267--283, 2018.
	
	\bibitem[Zheng et~al.(2020)Zheng, Gao, Liao, Hong, Zhang, Cui, and
	Chanussot]{zheng2020coupled}
	Ke~Zheng, Lianru Gao, Wenzhi Liao, Danfeng Hong, Bing Zhang, Ximin Cui, and
	Jocelyn Chanussot.
	\newblock Coupled convolutional neural network with adaptive response function
	learning for unsupervised hyperspectral super resolution.
	\newblock \emph{IEEE Transactions on Geoscience and Remote Sensing},
	59\penalty0 (3):\penalty0 2487--2502, 2020.
	
	\bibitem[Zhong et~al.(2020)Zhong, Bepler, Davis, and
	Berger]{zhong2020reconstructing}
	Ellen~D Zhong, Tristan Bepler, Joseph~H Davis, and Bonnie Berger.
	\newblock Reconstructing continuous distributions of 3d protein structure from
	cryo-em images.
	\newblock In \emph{International Conference on Learning Representations}, 2020.
	
	\bibitem[Zhong et~al.(2018)Zhong, Wang, Xu, Wang, Jia, Hu, Zhao, Wei, and
	Zhang]{Zhong2018}
	Yanfei Zhong, Xinyu Wang, Yao Xu, Shaoyu Wang, Tianyi Jia, Xin Hu, Ji~Zhao,
	Lifei Wei, and Liangpei Zhang.
	\newblock Mini-uav-borne hyperspectral remote sensing: From observation and
	processing to applications.
	\newblock \emph{IEEE Geoscience and Remote Sensing Magazine}, 6\penalty0
	(4):\penalty0 46--62, 2018.
	\newblock \doi{10.1109/MGRS.2018.2867592}.
	
	\bibitem[Zhu et~al.(2021)Zhu, Deng, Zheng, Zhong, Guan, Lin, Zhang, and
	Li]{zhu2021spectral}
	Qiqi Zhu, Weihuan Deng, Zhuo Zheng, Yanfei Zhong, Qingfeng Guan, Weihua Lin,
	Liangpei Zhang, and Deren Li.
	\newblock A spectral-spatial-dependent global learning framework for
	insufficient and imbalanced hyperspectral image classification.
	\newblock \emph{IEEE Transactions on Cybernetics}, 2021.
	
\end{thebibliography}

\bibliographystyle{iclr2024_conference}

\appendix
\newpage
\section{Appendix}
\label{sec:supplemental}

\subsection{A Illustration of Using \modelname~ for Multitask Image Super-Resolution} \label{sec:multitask_illus}
\begin{figure*}[ht!]
	\centering \tiny
	\vspace*{-0.2cm}
	\begin{subfigure}[b]{0.347\textwidth}  
		\centering 
		\includegraphics[width=\textwidth]{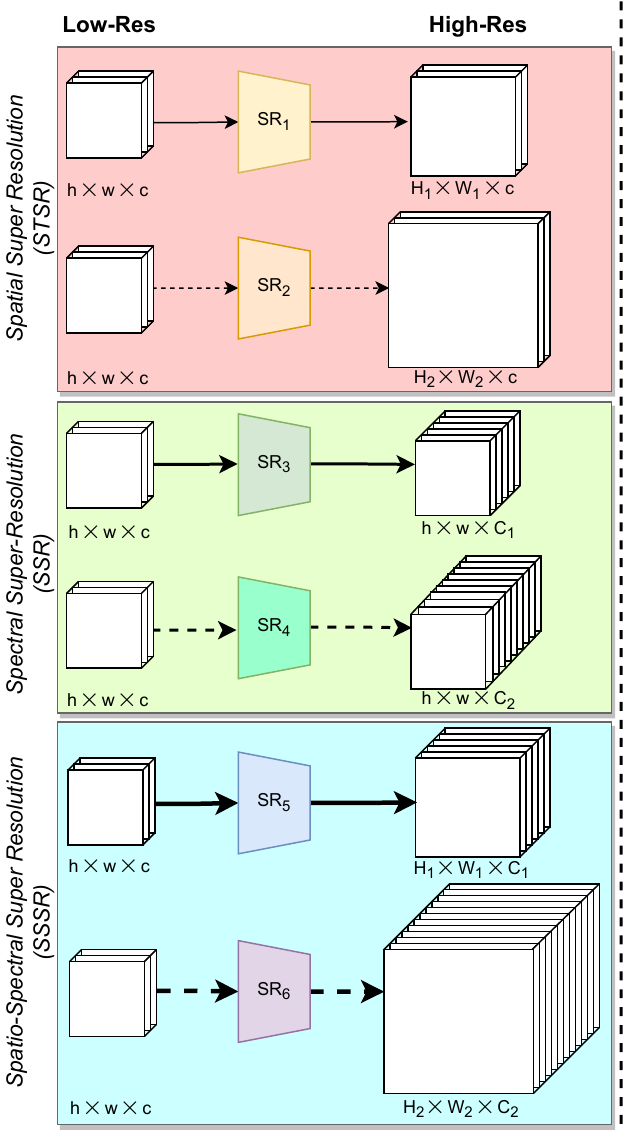}\vspace*{-0.2cm}
		\caption[]{{\small 
		Classic SR
		}}    
		\label{fig:moti-1}
	\end{subfigure}
	\begin{subfigure}[b]{0.317\textwidth}  
		\centering 
		\includegraphics[width=\textwidth]{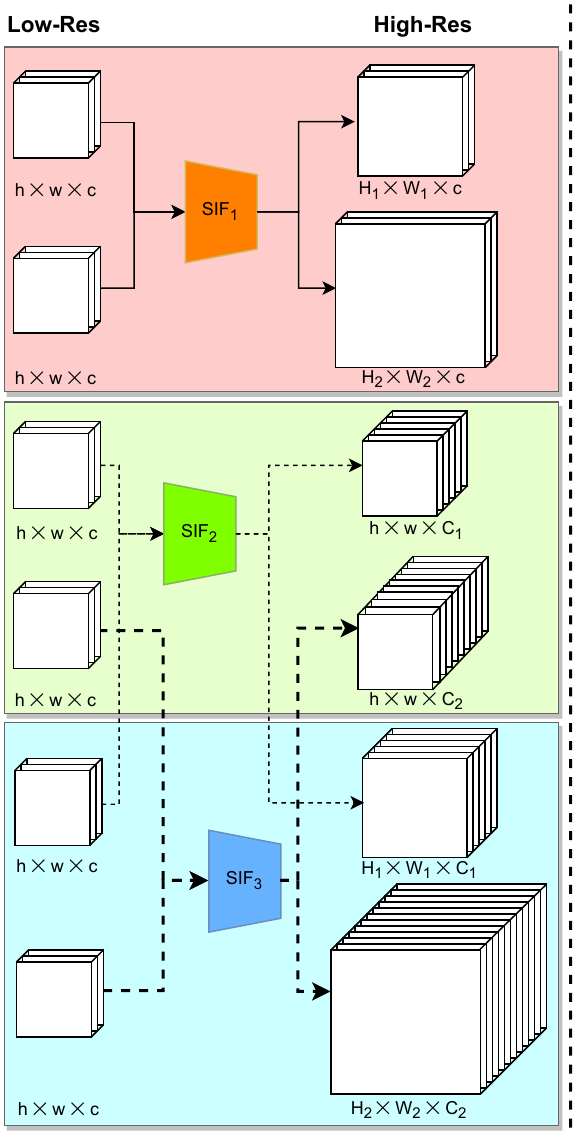}\vspace*{-0.2cm}
		\caption[]{{\small 
		Spatial Implicit Function
		}}    
		\label{fig:moti-2}
	\end{subfigure}
	\begin{subfigure}[b]{0.31\textwidth}  
		\centering 
		\includegraphics[width=\textwidth]{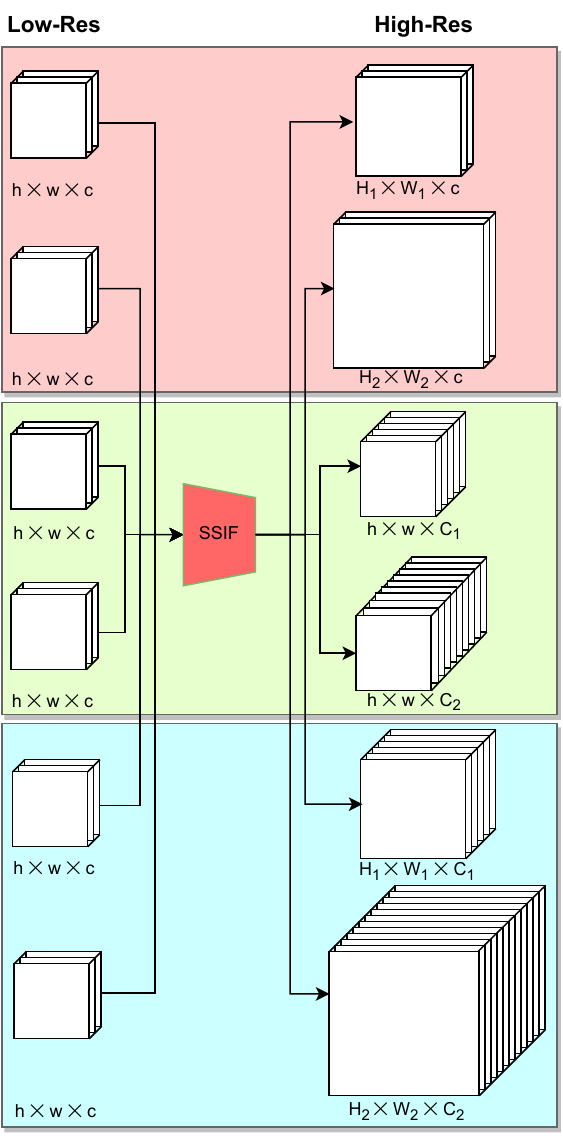}\vspace*{-0.2cm}
		\caption[]{{\small 
		\modelname
		}}    
		\label{fig:moti-3}
	\end{subfigure}
	
	\caption{
	An illustration of image super-resolution on different spatial and spectral resolutions. 
	The red, green, and blue boxes indicates three different super-resolution problems: Spatial Super-Resolution (spatial SR), Spectral Super-Resolution (spectral SR), and Spatio-Spectral Super-Resolution (SSSR). 
	The three subfigures illustrate how the classic super-resolution models, the spatial implicit functions, and \modelname~ handle different SR tasks which generated image with different spatial and spectral resolutions. 
(a) Classic SR - most super-resolution models train \textbf{separate SR models} for different input and output image pairs with different spatial and spectral resolutions such as RCAN \cite{zhang2018RCAN}, SR3\cite{saharia2021SR3}, SSJSR \cite{mei2020SepSSJSR}, \cite{he2021spatial};
	(b) Spatial Implicit Function (SIF) - recently many research focus on using the idea of neural implicit function to develop spatial scale-agnostic super-resolution models such that one model can be trained to do super-resolution for different spatial scale such as
	MetaSR\cite{hu2019MetaSR}, LIIF\cite{chen2021LIIF}, SADN \cite{wu2021SADN}, ITSRN \cite{yang2021ITSRN}, \cite{zhang2021implicit}. However, they have to train separate SR models if target images have different spectral resolutions.
	(c) Spatial-Spectral Implicit Function ($\modelname$) aims at using one model to handle different spatial scales and spectral scales at the same time such that we can train one generic model for different SR tasks. }
	\label{fig:motivation}
    \vspace*{-0.15cm}
\end{figure*}

\subsection{Spectral Encoder $\encband$} \label{sec:posenc}

A key component of $\modelname$ is the spectral encoder $\encband$ component. It  consists of a Fourier feature mapping layer $\psenc(\cdot)$ followed by a multi-layer perceptron $\specmlp(\cdot)$:
\begin{align}
\embband{i,\waveidx}  = \encband(\wave_{i,\waveidx}) = \specmlp(\psenc(\wave_{i,\waveidx}))
\label{eq:specenc_detail}
\end{align}

The Fourier feature mapping layer $\psenc(\cdot)$ takes a wavelength  $\wave_{i,\waveidx}$ sampled from the wavelength interval $\wavevecHRHSI_{i} = [\wave_{i,s}, \wave_{i,e}] \in \wavematHRHSI$ as the input and map it to a high dimensional vector 
$\embband{i,\waveidx} \in \Real^{\decdim}$, by using sinusoid functions with different frequencies. The idea is similar to the position encoder in Transformer \citep{vaswani2017attention}, NeRF \citep{mildenhall2020nerf}, Space2Vec \citep{mai2019space2vec,tancik2020fourier}, and 
spatial implicit functions \citep{zhang2021implicit,dupont2021coin} for pixel location encoding. Here, we adopt the Space2Vec \citep{mai2019space2vec} style position encoder $\psenc(\cdot)$. Let $\minscale, \maxscale$ be the minimum and maximum scaling factor in the wavelength space, and $\scalefac = \frac{\maxscale}{\minscale}$.
We define $\psenc(\cdot)$ as Equation \ref{eq:posencpt}). Here, $\bigcup_{\scalek=0}^{\nscale-1}$ indicates vector concatenation through different scales.
\vspace{-0.1cm}
\begin{align}
\psenc(\wave)=\bigcup_{\scalek=0}^{\nscale-1}\Big[\sin(\dfrac{\wave}{\minscale \cdot \scalefac^{\scalek/(\nscale-1)}}),\cos(\dfrac{\wave}{\minscale \cdot \scalefac^{\scalek/(\nscale-1)}})\Big]; 
\label{eq:posencpt}
\end{align}

\subsection{Super-Resolution Data Preparation} \label{sec:data_perp_app}

Figure \ref{fig:model_data_prep} illustrates the data preparation process of \modelname.
Given a training image pair which consists of a high spatial-spectral resolution image $\imgHRHSIMAX \in \Real^{\hhr \times \whr \times \cmax}$ and an image with high spatial resolution but low spectral resolution $\imgHRMSI \in \Real^{\hhr \times \whr \times \clr}$, we perform downsampling in both the spectral domain and spatial domain.

For the spectral downsampling process (the blue box in Figure \ref{fig:model_data_prep}), we randomly sample a band number $\chr \sim \uniformdist(\cmin, \cmax)$ from a uniform distribution between the minimum and maximum band number $\cmin, \cmax > 0$.
We use $\chr$ to downsample $\imgHRHSIMAX$ in the spectral domain which yield $\imgHRHSI \in \Real^{\hhr \times \whr \times \chr}$. 
Then we convert $\imgHRHSI$ into location-value-wavelength samples $(\ptHRHSI{}, \ptvalHRHSI{}, \wavevecHRHSI)$. 
$\ptHRHSI{}$ and $\wavevecHRHSI$ serve as the input features while $\ptvalHRHSI{}$ are the prediction target. Note that, here we can sample equally spaced wavelength intervals or irregular spaced wavelength intervals for the target HR-HSI images $\imgHRHSI$ since \modelname~ is agnostic to this irregularity.

For the spatial downsampling (the orange box in Figure \ref{fig:model_arch}), we randomly sample a spatial scale $\scalespa \sim \uniformdist(\scalespaMIN, \scalespaMAX)$ where $\uniformdist(\scalespaMIN, \scalespaMAX)$ is a uniform distribution between the minimum and maximum spatial scale $\scalespaMIN, \scalespaMAX > 0$.
We use $\scalespa$ to spatially downsample $\imgHRMSI$ into $\imgLRMSI \in \Real^{\hlr \times \wlr \times \clr}$ which serves as the input for $\modelname$. Here, $\hlr = \hhr/\scalespa$ and $\wlr = \whr/\scalespa$.

Interestingly, when the spatial upsampling scale $\scalespa$ is fixed as 1, our \modelname~ is degraded to a spectral SR model. When the band $\chr$ is fixed as the same as the input band, i.e., $\chr = \clr$, \modelname~ is degraded to a spatial SR model. When we vary $\chr$ and $\scalespa$ during \modelname~ training, we allow the model to do spatial SR and spectral SR at different difficulty levels which helps it to learn a continuous representation both in the spatial and spectral domain. 
\subsection{Baslines} \label{sec:baseline_app}
We consider 7 baselines in our SSSR task on two benchmark dataset: 
\begin{enumerate}[leftmargin=0.cm]
\setlength\itemsep{0em}
    \item \textbf{RCAN + AWAN} uses RCAN \citep{zhang2018RCAN} for spatial SR and then  AWAN \citep{li2020AWAN} for spectral SR in a sequential manner.
    \item \textbf{AWAN + RCAN} simply reverses the order of RCAN and AWAN.
    \item \textbf{AWAN + SSPSR} uses AWAN and SSPSR \citep{mei2020SepSSJSR} for spectral SR and spatial SR.
    \item \textbf{RC/AW + MoG-DCN} first separately uses RCAN \citep{zhang2018RCAN} to do spatial SR to obtain HR-MSI images and 
    uses AWAN \citep{li2020AWAN} to do spectral SR to obtain LR-HSI images, and then uses MoG-DCN \citep{dong2021MoGDCN} to do hyperspectral image fusion based on the previously generated HR-MSI and LR-HSI images.
    \item \textbf{SSJSR} \citep{mei2020SepSSJSR} uses a fully convolution-based deep neural network to do SSSR.
    \item \textbf{US3RN} \citep{ma2021US3RN} uses a deep unfolding network to solve the SSSR problem with a close-form solution. It is the current state-of-the-art model for the SSSR task.
    \item \textbf{LIIF} \citep{chen2021LIIF} is initially designed for spatial SR on multispectral data. We increase the output dimension of LIIF's final MLP to allow it to work on hyperspectral images.
\end{enumerate} \subsection{Dataset Description}  \label{sec:data}
The CAVE dataset \citep{yasuma2010generalized} consists of 32 indoor hyperspectral (HSI) images captured under controlled illumination. Each image has a spatial size of $512 \times 512$ and 31 spectral bands covering the wavelength from 400nm to 700nm. Each HSI image is associated with an RGB image with the same spatial size. There are a lot of study using the CAVE dataset for hyperspectral image super-resolution
\citep{yao2020cross,mei2020SepSSJSR,zhang2020unsupervised,zhang2021implicit,han2021spectral,qu2021unsupervised,ma2021US3RN}. 
However, these works focus on different SR tasks.
In this work, we focus on the most challenging task -- SSSR. The train/test split on the CAVE dataset varies from paper to paper. In order to keep a fair comparison to the previous study, we adopt the train/test split from US3RN \citep{ma2021US3RN}, the lastest work on this dataset, and use the first 22 samples as the training dataset and the rest 10 samples as testing. The limited number of samples pose a significant challenges on modeling training. So similar to the previous work \citep{ma2021US3RN,chen2021LIIF}, given a HR-HSI and HR-MSI image pair $(\imgHRHSIMAX, \imgHRMSI)$, we first do random croping for a $64\scalespa \times 64 \scalespa $ image patch from both images. Then $\imgHRMSI$ is spatially downsampled to a $64 \times 64$ image patch which serves as the input LR-MSI image $\imgLRMSI$. 
We choose $\scalespaMIN = 1$ and $\scalespaMAX = 8$ and $\scalespaMIN = 1$ for spatial downsampling,  $\cmin = 8$ and $\cmax = 31$ for spectral downsampling (See Section \ref{sec:data_prep}).

The Pavia Centre (PC) dataset is taken by ROSIS, a widely used hyperspectral sensor. The images were collected over the center area of Pavia, northern Italy, in 2001. It contains 102 spectral bands covering secptrum from 430nm to 860nm. Figure \ref{fig:ssif_cont} shows the spectral signature of one pixel A when $\chr = 102$. It has $1095 \times 715$ effective pixels. Similarly, we also adopt the train/test split from US3RN \citep{ma2021US3RN} and crop the upper left $1024 \times 128$ pixels as the testing dataset and the rest for training. The PC dataset does not come with a multispectral image counterpart. So we adopt the practice of \citep{mei2020SepSSJSR} to simulate the high-resolution multispectral (HR-MSI) image based on the sensor specification of the multispectral sensor of IKONOS. The resulted image has 4 bands which correspond to R, G, B, and NIR. Please see the MSI spectral signature in Figure \ref{fig:ssif_cont} for reference. Same random cropping technique is used for PC. We choose $\scalespaMIN = 1$ and $\scalespaMAX = 8$ and $\scalespaMIN = 1$ for spatial downsampling,  $\cmin = 13$ and $\cmax = 102$ for spectral downsampling (See Section \ref{sec:data_prep}). 
\subsection{\modelname~ Implementation and Training Details} \label{sec:training_detail_app}

We use RDN\citep{zhang2018RDN} as the image encoder $\encimg(\cdot)$ and we use LIIF \citep{chen2021LIIF} as the pixel feature decoder $\decsif(\cdot)$.

For both CAVE and Pavia Centre dataset, we first tune the learning rate $lr = \{5.e-5, 1.e-4, 2.e-4\}$. We find out the default learning rate used by LIIF $lr = 1.e-4\}$ works the best for both datasets.

Then we tune the hyperparameters of LIIF
including the output image feature dimension for image encoder $\encimg(\cdot)$ -- $\encimgcdim = \{64, 128, 256\}$, the input image size $\hlr = \wlr \in \{48, 64\}$, the hidden dimension of LIIF's multi-layer perceptron -- $\liifmlphiddim \in \{256, 512\}$. 
We find out  $\encimgcdim = 64$, $\hlr = \wlr = 64$, and $\liifmlphiddim=256$ give us the best results of LIIF on CAVE while for the Pavia Centre, $\encimgcdim = 256$, $\hlr = \wlr = 64$, and $\liifmlphiddim=512$ yield the best results. In addition, we find out using multiple dataloaders with different input image sizes $\hlr = \wlr$ is especially useful for the Pavia Centre dataset. In our experiment, we use three different dataloaders with $\{16, 32, 64\}$ as their input image size.

After we get the best hyperparameter combination of LIIF, we directly use them for \modelname without tuning. And we only tune the newly added hyperparameters for \modelname including the hidden dimension $\ssifmlphiddim = \{512, 1024\}$ of $\specmlp(\cdot)$ in Equation \ref{eq:specenc_detail}
and the wavelength sampling number $\numsampwave \in \{2,4,8,16,32,48,52,64\}$. We find out $\ssifmlphiddim = 512$ and $\numsampwave = 16$ are the best hyperparameter combination for the CAVE dataset and $\ssifmlphiddim = 1024$ and $\numsampwave = 64$ is the best for the Pavia Centre dataset.

All experiments are conducted on a Linux server with 1 CUDA GPU of 12GB memory. We use the official implementations\footnote{The LIIF implementation is under BSD 3-Clause "New" or "Revised" License.} of LIIF \citep{chen2021LIIF} and US3RN \citep{ma2021US3RN}. We implement our \modelname~ in PyTorch and will be made publicly available upon acceptance. 

\subsection{Additional Experiment Results on the CAVE dataset} \label{sec:cave_res_app}

\begin{figure*}[t!]
	\centering \tiny
	\vspace*{-0.2cm}
	\begin{subfigure}[b]{0.400\textwidth}  
		\centering 
		\includegraphics[width=\textwidth]{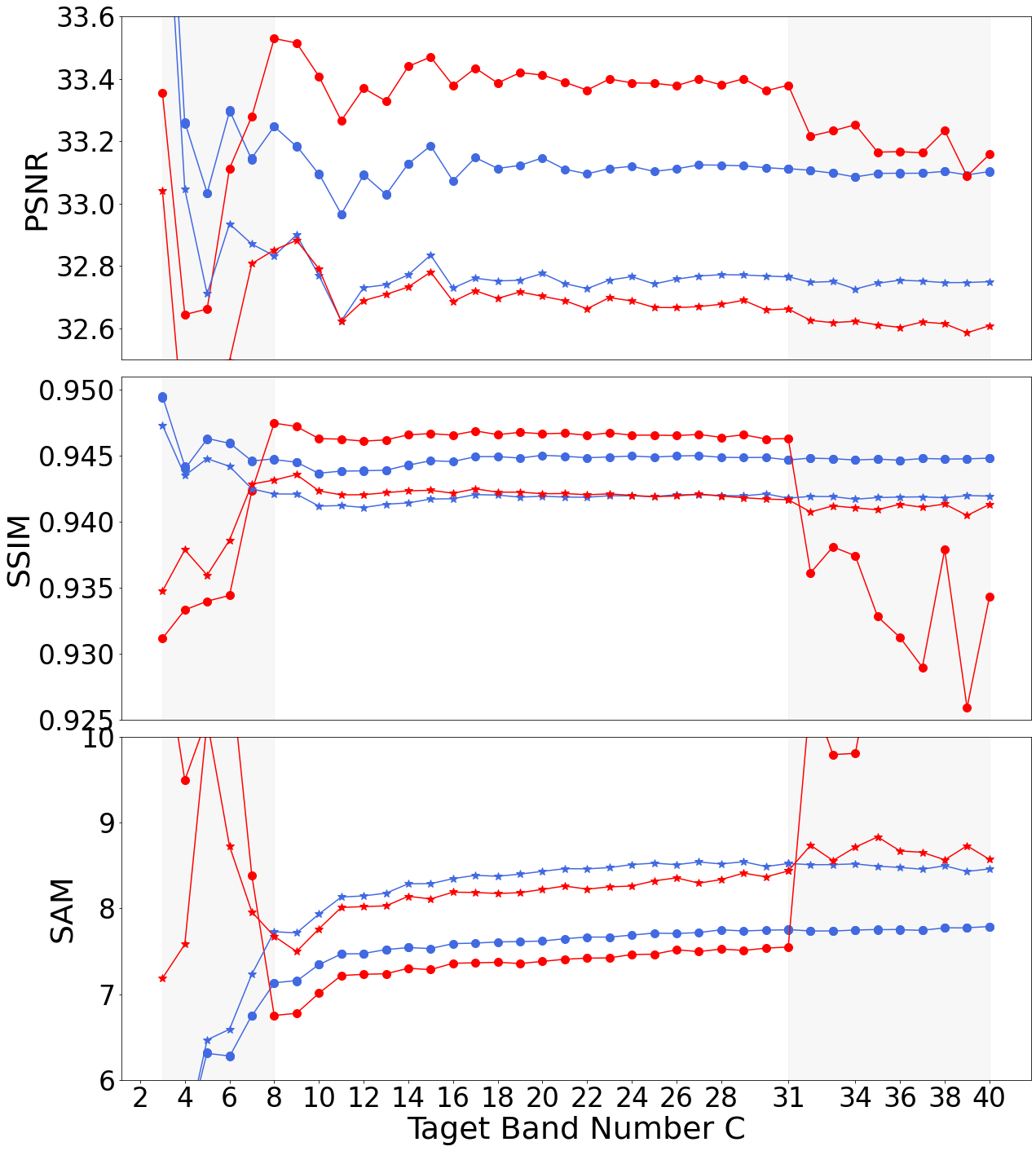}\vspace*{-0.2cm}
		\caption[]{{\small 
		Scale $\scalespa = 4$
		}}    
		\label{fig:cave_eval_scale4_abla_CD}
	\end{subfigure}
 	\hspace{0.05\textwidth}
	\begin{subfigure}[b]{0.535\textwidth}  
		\centering 
		\includegraphics[width=\textwidth]{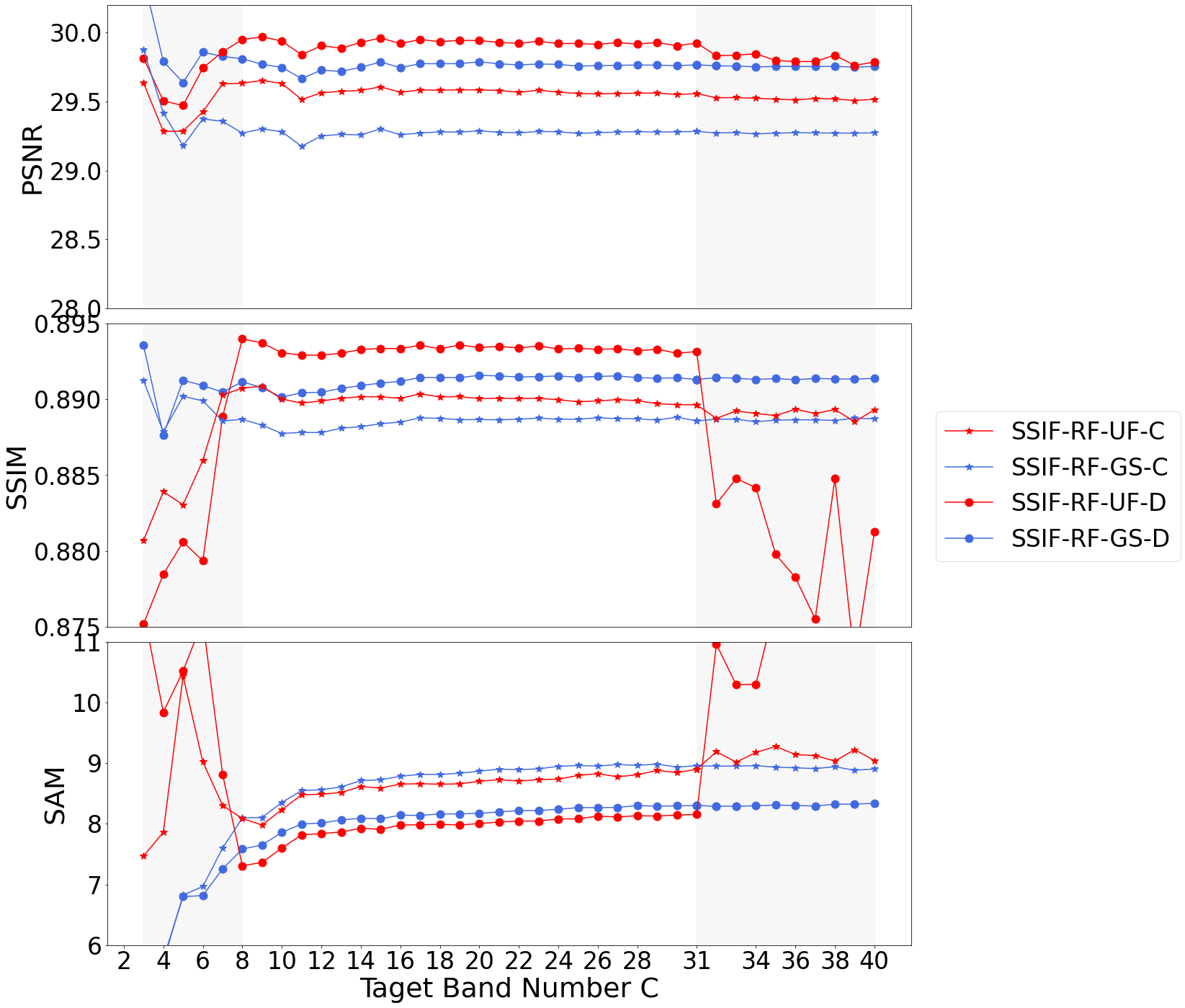}\vspace*{-0.2cm}
		\caption[]{{\small 
		Scale $\scalespa = 8$
		}}    
		\label{fig:cave_eval_scale8_abla_CD}
	\end{subfigure}
 \vspace*{-0.4cm}
	\caption{
	The ablation studies of different designs of spectral decoder $\decband$ on the CAVE dataset. Here, we use two \modelname{} models -- \modelname-RF-GS (blue curves) and \modelname-RF-UF (red curves). Two spectral decoder $\decband$ variants are explored: ``\textbf{D}'' and ``\textbf{C}''.
	}
	\label{fig:cave_eval_abla_CD}
    \vspace*{-0.2cm}
\end{figure*}

Figure \ref{fig:cave_eval_abla_CD} illustrates the results of our ablation studies on different designs of spectral decoder $\decband$ on the CAVE dataset. Two \modelname{} models -- \modelname-RF-GS and \modelname-RF-UF -- are used. We test two spectral decoder $\decband$ variants:
\begin{enumerate}
    \item ``\textbf{D}'': $\decband$ is a multilayer perception (MLP) which is modulated by the image feature embedding $\embband{i,\waveidx}$. $\decband$ takes a spectral embedding $\embband{i,\waveidx}$ as the input and output the corresponding radiance value. When $\decband$ is a one-layer MLP, this can be seen as the dot product between the input spectral embedding $\embband{i,\waveidx}$ and image feature embedding $\embband{i,\waveidx}$.
    \item ``\textbf{C}'': $\decband$ is a multilayer perception (MLP) which takes the concatenation of spectral embedding $\embband{i,\waveidx}$ and image feature embedding $\embband{i,\waveidx}$ as the input and output the corresponding radiance value.
\end{enumerate}

Two \modelname{} models and two spectral decoder $\decband$ variants amount to 4 different \modelname{} variants. From Figure  \ref{fig:cave_eval_abla_CD}, we can see that:
\begin{enumerate}
    \item \modelname-RF-*-D usually outperform \modelname-RF-*-C which indicates that spectral decoder $\decband$ variant \textbf{D} is usually more effective than \textbf{C}.
    \item We find out \modelname-RF-UF-D can outperform \modelname-RF-UF-C for in-distribution spectral resolutions (i.e., $ 8 \leq \chr \leq 31 $) while \modelname-RF-UF-C has better generalizability than \modelname-RF-UF-D for out-of-distribution spectral resolutions (i.e., $ \chr > 31 $).
\end{enumerate} 
\subsection{Additional Experiments Results on the Pavia Centre dataset} \label{sec:pc_res_app}

\begin{figure*}[ht!]
	\centering 
	\includegraphics[width=0.95\textwidth]{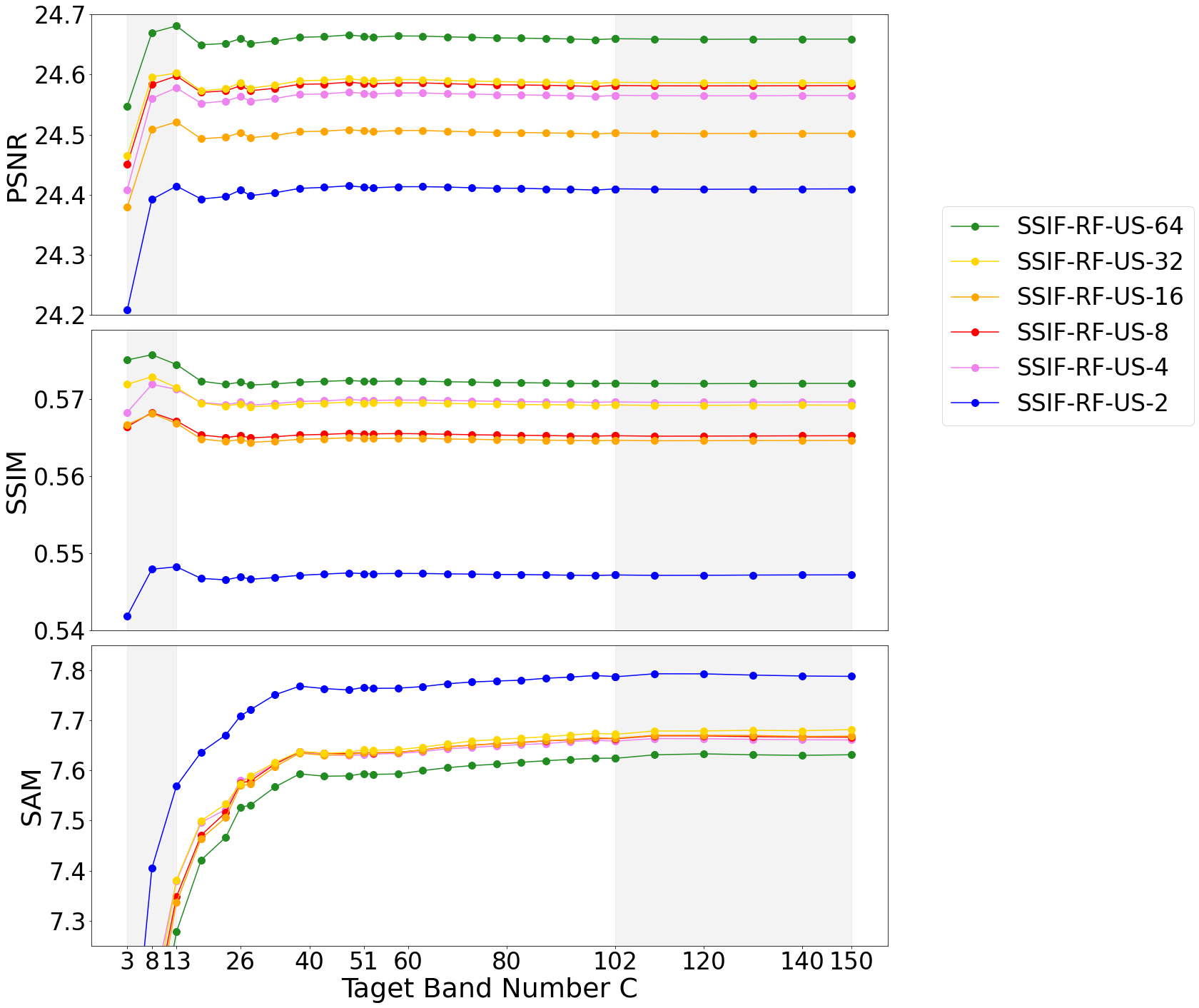}\vspace*{-0.2cm}
	\caption{
	The ablation study on the number of sampled wavelengths in each wavelength interval $\wavevecHRHSI_{i}$ -- $\numsampwave$ on the Pavia Centra dataset with spatial scale $\scalespa = 8$. The setting is similar to Figure \ref{fig:pc_eval_band}. We use \modelname-RF-US model as an example and tune the hyperparameter $\numsampwave = \{2,4,6,8,16,32,64\}$. Here, each \modelname{} is named as \modelname-RF-US-\numsampwave. We can see that a bigger $\numsampwave$ leads to better model performance.
	} 
	\label{fig:pc_ablation_k}
	\vspace*{-0.15cm}
\end{figure*}

We conduct the ablation study on the effect of the number of sampled wavelengths in each wavelength interval $\wavevecHRHSI_{i}$ -- $\numsampwave$ on the model performance. We use the Pavia Centra dataset as an example and compare model performances of \modelname-RF-US with different $\numsampwave$. Figure \ref{fig:pc_ablation_k} illustrates the results. We can see that a bigger $\numsampwave$ leads to better model performance.

\begin{figure*}[t!]
	\centering \tiny
	\vspace*{-0.2cm}
	\begin{subfigure}[b]{1.0\textwidth}  
		\centering 
		\includegraphics[width=\textwidth]{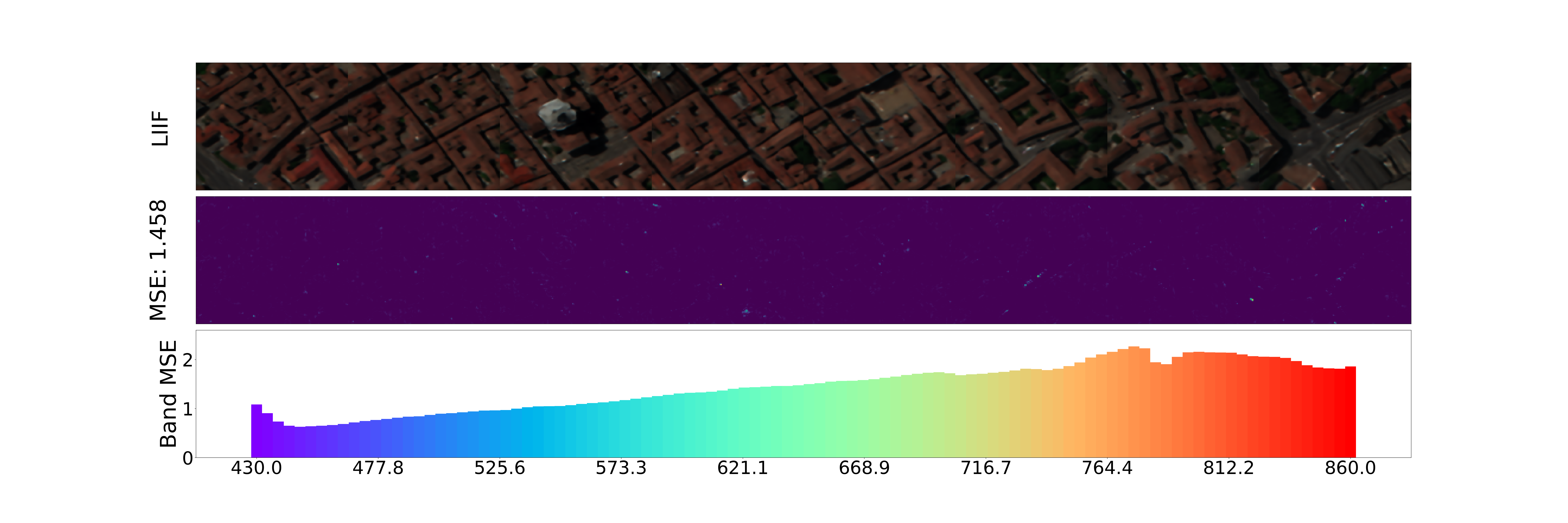}\vspace*{-0.2cm}
		\caption[]{{\small 
		LIIF
		}}    
		\label{fig:liif}
	\end{subfigure}
\begin{subfigure}[b]{1.0\textwidth}  
		\centering 
		\includegraphics[width=\textwidth]{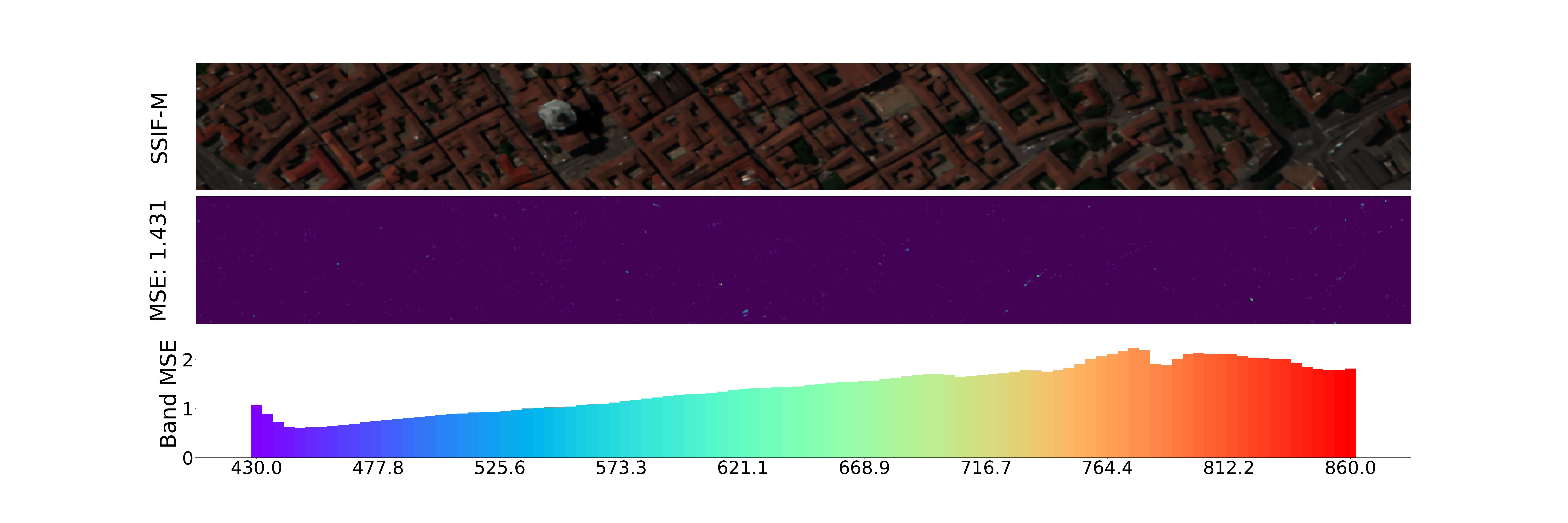}\vspace*{-0.2cm}
		\caption[]{{\small 
		SSIF-M
		}}    
		\label{fig:ssif-m}
	\end{subfigure}
	\begin{subfigure}[b]{1.0\textwidth}  
		\centering 
		\includegraphics[width=\textwidth]{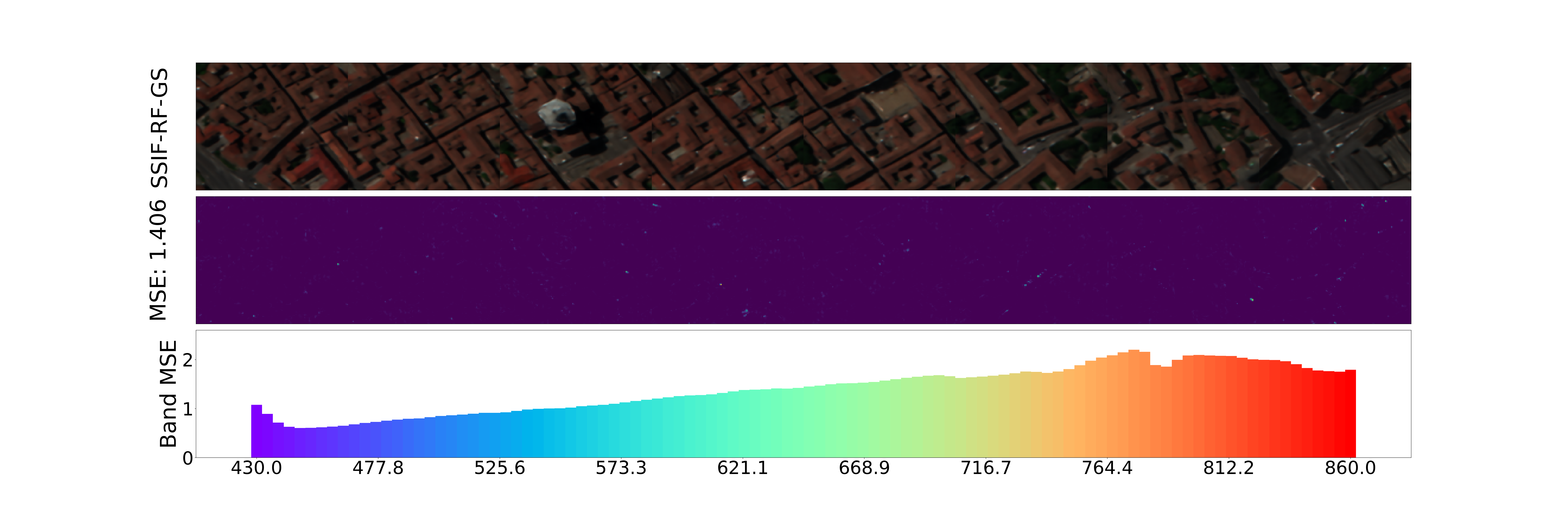}\vspace*{-0.2cm}
		\caption[]{{\small 
		SSIF-RF-GS
		}}    
		\label{fig:ssif-rf-gs}
	\end{subfigure}
	\begin{subfigure}[b]{1.0\textwidth}  
		\centering 
		\includegraphics[width=\textwidth]{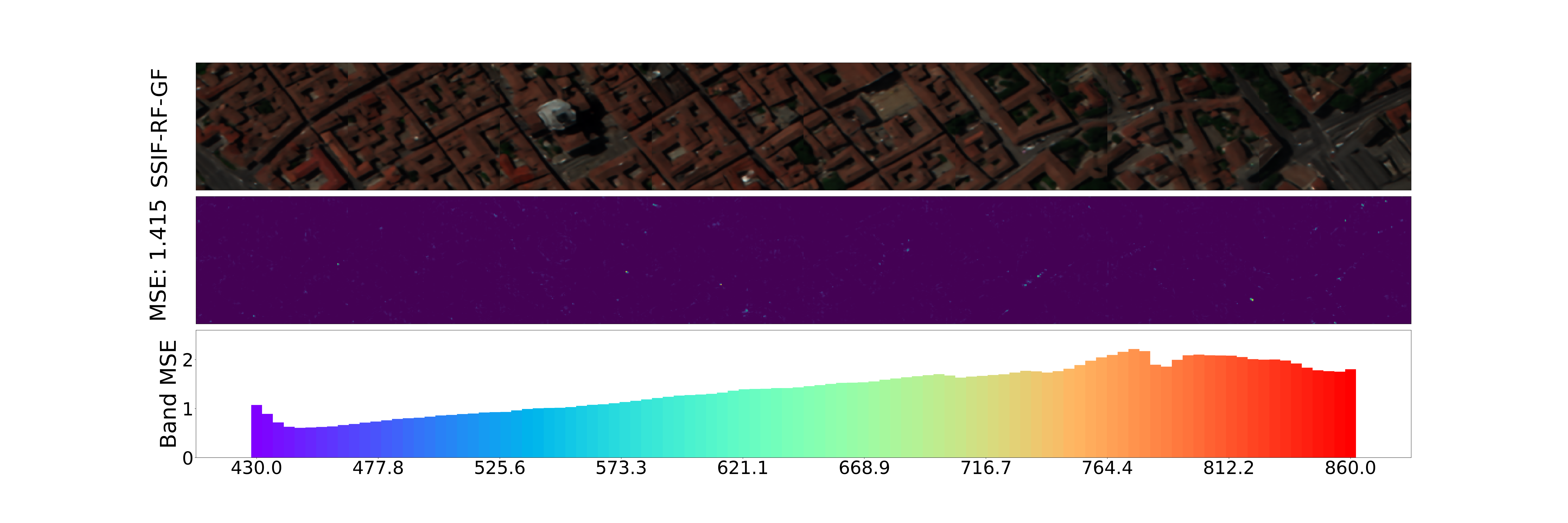}\vspace*{-0.2cm}
		\caption[]{{\small 
		SSIF-RF-GF
		}}    
		\label{fig:ssif-rf-gf}
	\end{subfigure}
	\caption{
	The comparison among the generated images from LIIF and different \modelname~ for $\scalespa=4$ and $\chr = 102$. For Figure (a)-(g), we first show the generated image of the corresponding model. Then we show the MSE between it and the ground truth image per pixel. The y axis label indicates the MSE value which times 1000.  Then finally, we show the MSE for each band.
	Figure (g) shows the ground truth image.
	}
	\label{fig:generate_img}
    \vspace*{-0.15cm}
\end{figure*}

\begin{figure*}[t!]\ContinuedFloat
	\centering \tiny
	\vspace*{-0.2cm}
	\begin{subfigure}[b]{1.0\textwidth}  
		\centering 
		\includegraphics[width=\textwidth]{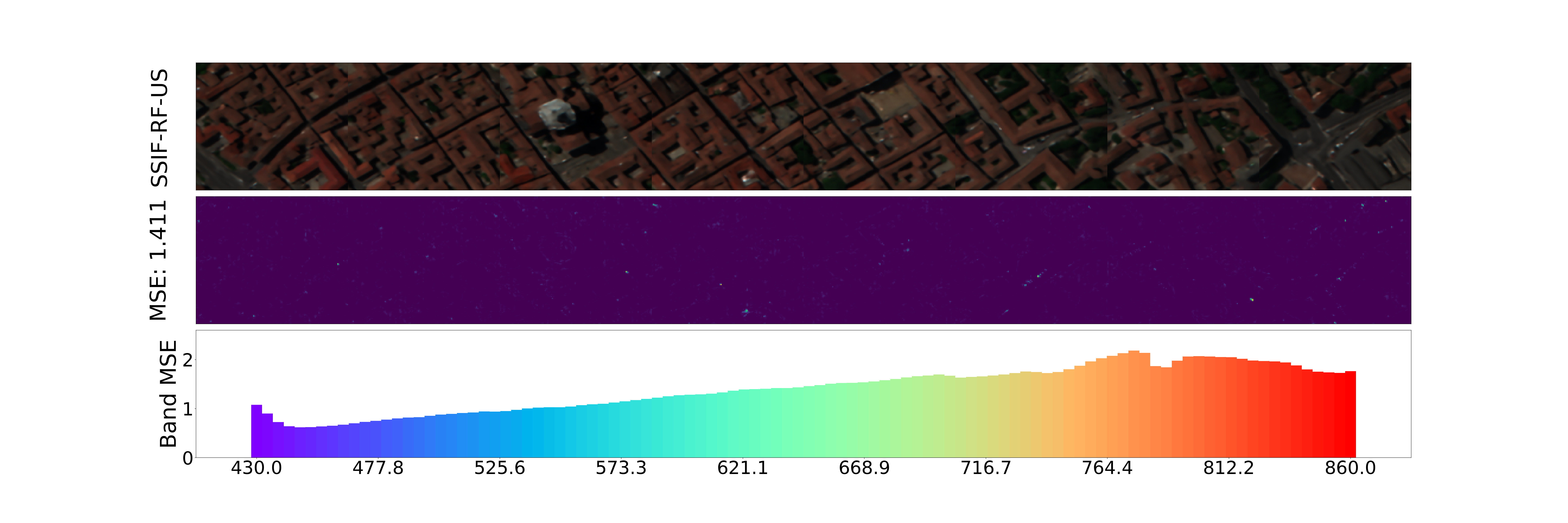}\vspace*{-0.2cm}
		\caption[]{{\small 
		SSIF-RF-US
		}}    
		\label{fig:ssif-rf-us}
	\end{subfigure}
        \begin{subfigure}[b]{1.0\textwidth}  
		\centering 
		\includegraphics[width=\textwidth]{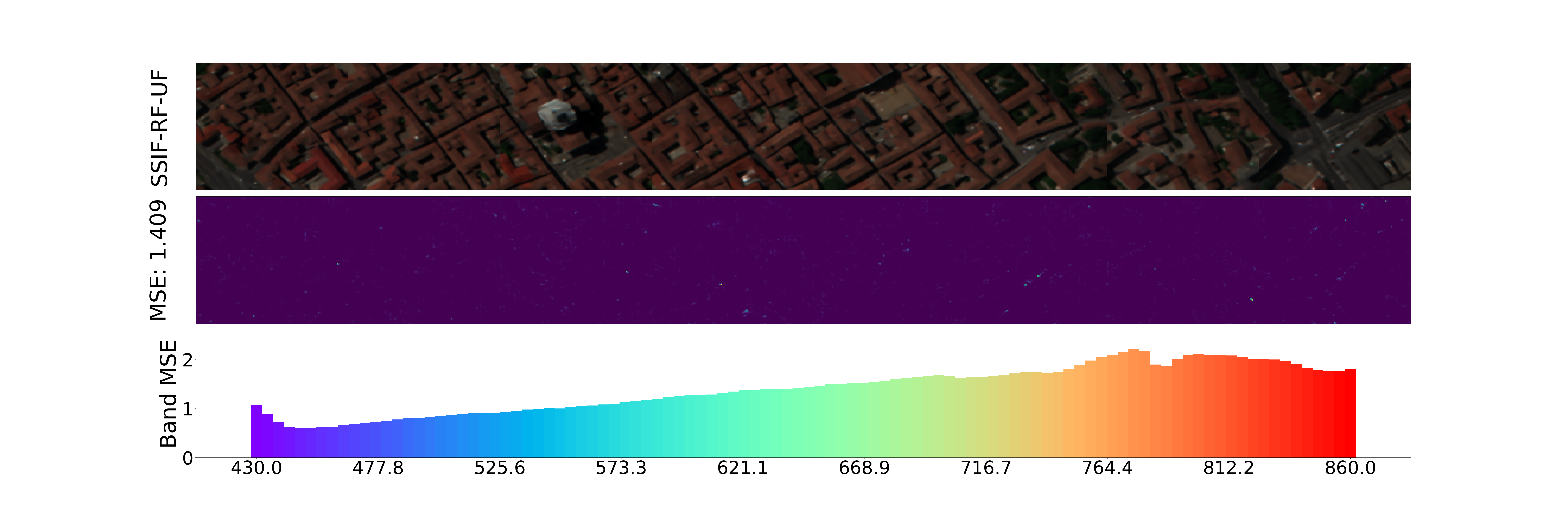}\vspace*{-0.2cm}
		\caption[]{{\small 
		SSIF-RF-UF
		}}    
		\label{fig:ssif-rf-uf}
	\end{subfigure}
	\begin{subfigure}[b]{1.0\textwidth}  
		\centering 
		\includegraphics[width=0.81\textwidth]{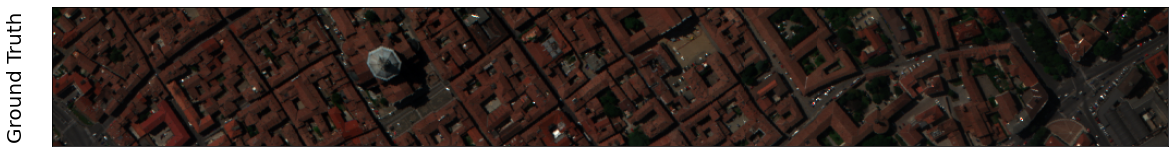}\vspace*{-0.2cm}
		\caption[]{{\small 
		Ground Truth
		}}    
		\label{fig:img_gt}
	\end{subfigure}
	\caption{
	The comparison among the generated images from LIIF and different \modelname~ for $\scalespa=4$ and $\chr = 102$. For Figure (a)-(g), we first show the generated image of the corresponding model. Then we show the MSE between it and the ground truth image per pixel. The y axis label indicates the MSE value which times 1000.  Then finally, we show the MSE for each band.
	Figure (g) shows the ground truth image.
	}
	\label{fig:generate_img}
    \vspace*{-0.15cm}
\end{figure*}

\subsection{Descriptions of the Land Use Classification Task on Pavia Centre Dataset} \label{sec:lulcc-app}

\paragraph{Motivation}
In addition to directly comparing the generated images with the ground truth images by using those image similarity metrics, many super-resolution works use human evaluation to evaluate whether the generated image looks natural or not \citep{khrulkov2021neural,saharia2021SR3,he2021spatial}. However, as for super-resolution on remote sensing (RS) image datasets like Pavia Centre, the objective is not to generate RS image that look natural for human eyes but to generate RS images which can be useful for downstream tasks such as land use classification.

\paragraph{Pavia Centre Land Use Classification Dataset}
So in this work, we conduct an additional evaluation on our \modelname~ as well as the strongest baseline -- LIIF by using land use classification task to test the fidlity of the generated RS images.
The Pavia Centre dataset comes with a human annotated land use classification map with 10 different land use types for each pixel including water, trees, asphalt, self-blocking bricks, bitumen, tiles, shadows, meadows, and bare soil. For a statistic for each land use types, please refer to the original Pavia Centre webpage\footnote{\url{http://www.ehu.eus/ccwintco/index.php/Hyperspectral_Remote_Sensing_Scenes}}. So we use it as the ground truth labels. 

\paragraph{Land Use Classification Model}
In terms of the land use classification model, we need to select an appropriate image segmentation model for the Pavia Centre dataset. The recent Segmenter \citep{strudel2021segmenter} network extends the Vision Transformer to a semantic segmentation model which shows the state-of-the-art performance on multiple RGB image segmentation datasets. \cite{ayush2021geography} utilized PSANet \citep{zhao2018psanet} network with ResNet-50 backbone to perform semantic segmentation on the SpaceNet satellite image dataset \citep{van2018spacenet}. However, both models are designed for semantic segmentation on RBG images while what we want is an image segmentation model on hyperspectral images which contain hundreds of bands. 
SSJSR \citep{mei2020SepSSJSR} chooses a simple per-pixel-based support vector machine (SVM) model to do image segmentation on hyperspectral RS images for the super-resolution model evaluation. However, SVM does not consider the spatial neighborhood of each pixel so it cannot take into account the spatial correlations among nearby pixels which will lead to suboptimal results. We finally choose to use A2S2K-ResNet \citep{roy2020A2S2K} as our image segmentation model. 
A2S2K-ResNet is ranked the third place in the PaperWithCode leaderboard\footnote{\url{https://paperswithcode.com/sota/hyperspectral-image-classification-on-pavia}} on the Pavia University dataset which is a hyperspectral image classification/segmentation dataset. The hyperspectral images from the Pavia University dataset was taken by exactly the same ROSIS sensors as the Pavia Centre dataset. Moreover, both datasets were collected at relatively the same time and nearby locations. Since we cannot find a leaderboard for the Pavia Centre dataset, we choose to use the leaderboard of the Pavia University dataset for reference. Note that the first ranked model -- SpectralNET \citep{chakraborty2021spectralnet} is from an unpublished ArXiv paper. And both the first and second-ranked model -- SpectralNET \citep{chakraborty2021spectralnet} and 	
SSDGL \citep{zhu2021spectral} have a poorly organized codebase which causes difficulties for us to reproduce their results. So we choose the third-ranked model -- A2S2K-ResNet \citep{roy2020A2S2K} which is a modified ResNet model for hypersepectral image classification/segmentation.

\paragraph{Model Training Detail}
We follow the exact training process of A2S2K-ResNet. Within the training region of the Pavia Centre hyperspectral image, we balanced sample 2000 pixel samples for each land use type. For each training pixel, we crop a $11 \times 11$ spatial neighborhood region as the input for A2S2K-ResNet. The model takes the $11 \times 11 \times 102$ input tensor and produces a probability distribution over all land use types to the current center pixel. In the test region, we also balanced sample 500 pixel samples for each land use type as the validation dataset. As reported in the \cite{roy2020A2S2K}, We train A2S2K-ResNet for 200 epochs and take the one model instance that has the highest performance on our validation dataset. We use this trained A2S2K-ResNet model to produce the predicted land use classification maps based on either the ground truth HSI images or the generated images from LIIF or \modelname~ for the test region. And then, we obtain the classification accuracy on each image. The results are shown in Table \ref{tab:lulcc_c102_short}.

\begin{figure*}[ht!]
	\centering 
	\includegraphics[width=0.95\textwidth]{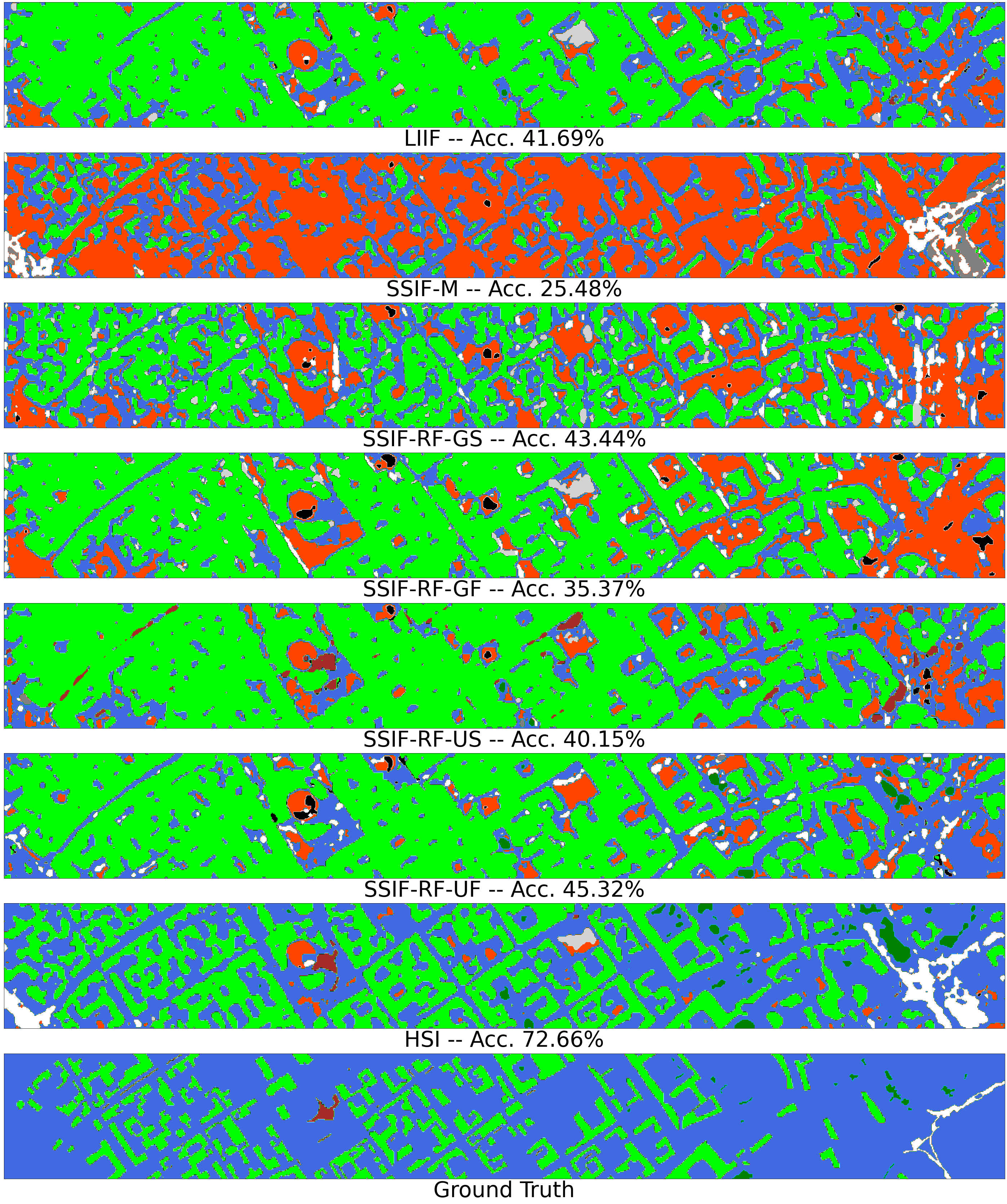}\vspace*{-0.2cm}
	\caption{
	The land use classification results of A2S2K-ResNet \citep{roy2020A2S2K} on the generated images under spatial upsampling scale $\scalespa=2$ from LIIF, \modelname-M, \modelname-RF-GS, \modelname-RF-GF, \modelname-RF-US, \modelname-RF-UF, as well as on the original Pavia Centre test image. "Ground Truth" indicates the ground truth labels. The classification accuracy is also listed under each figure.
	} 
	\label{fig:lulcc_cla_res2}
	\vspace*{-0.15cm}
\end{figure*} 
\subsection{Visualizing The Basis in Spectral Encoder} \label{sec:spec-basis}

\begin{figure*}[t!]
	\centering \tiny
	\vspace*{-0.2cm}
	\begin{subfigure}[b]{0.690\textwidth}  
		\centering 
		\includegraphics[width=\textwidth]{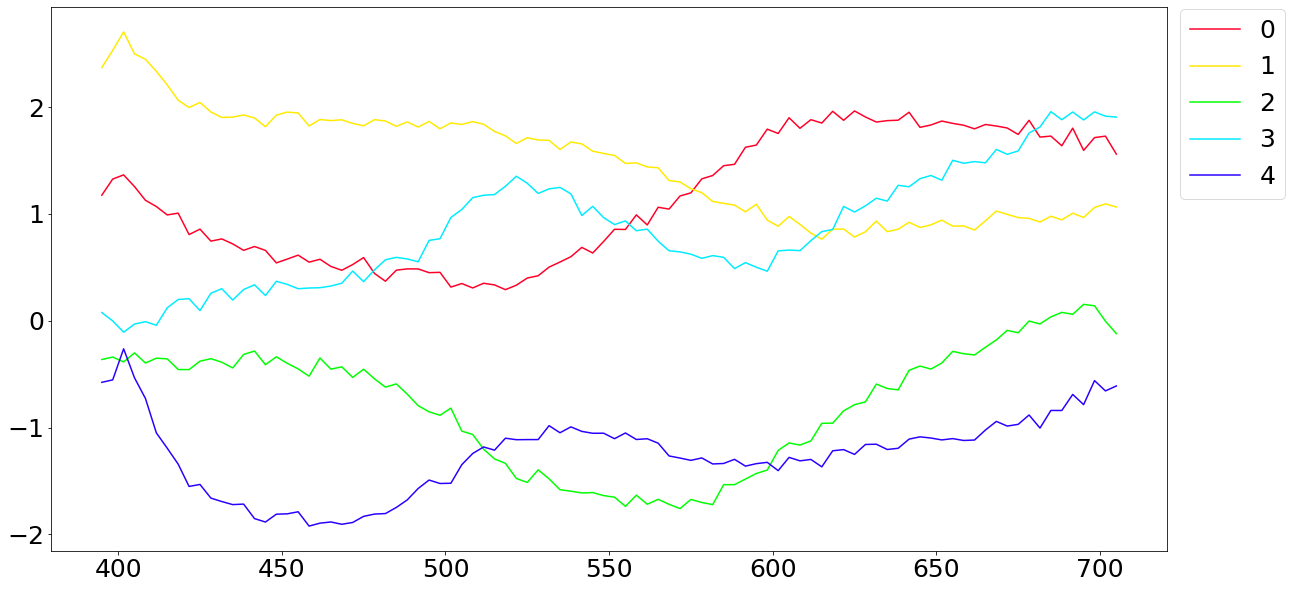}\vspace*{-0.2cm}
		\caption[]{{\small 
		Spectral embedding Dimension $\decdim = 5$
		}}    
		\label{fig:cave_specenc_dim5}
	\end{subfigure}
\hspace{0.05\textwidth}
	\begin{subfigure}[b]{0.690\textwidth}  
		\centering 
		\includegraphics[width=\textwidth]{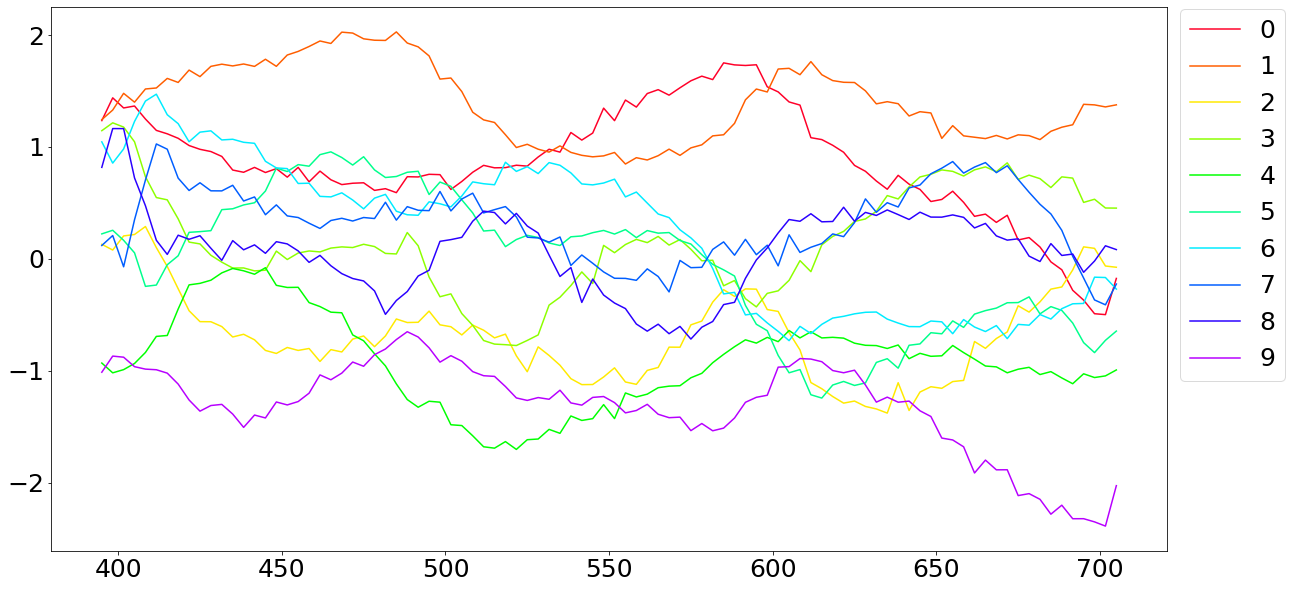}\vspace*{-0.2cm}
		\caption[]{{\small 
		Spectral embedding Dimension $\decdim = 10$
		}}    
		\label{fig:cave_specenc_dim10}
	\end{subfigure}
 \vspace*{-0.2cm}
	\caption{
	Visualizations of the spectral embeddings  with small spectral embedding dimensions $\decdim = \{5, 10\}$. Here we draw a curve for each dimension of the embedding, derived from the spectral encoders $\encband$ of two learned \modelname-RF-GS. The x axis indicates the wavelength and each curve $\encband(\wave){}[j]$ corresponds to the values of a specific spectral embedding dimension $j$.
	}
	\label{fig:cave_specenc_dim}
    \vspace*{-0.3cm}
\end{figure*}

\end{document}